\documentclass[11pt]{article}
\usepackage[final]{acl}
\usepackage{times}
\usepackage{latexsym}
\usepackage[T1]{fontenc}
\usepackage[utf8]{inputenc}
\usepackage{microtype}
\usepackage{inconsolata}
\usepackage{graphicx}
\usepackage{amsmath}
\usepackage{amssymb}
\usepackage{algorithm}
\usepackage{algpseudocode}
\usepackage{booktabs}
\usepackage{multirow}
\usepackage{xcolor}
\usepackage{placeins}
\usepackage{subcaption}
\usepackage{caption}

\title{Asymmetric On-Policy Distillation: Bridging Exploitation and \\Imitation at the Token Level}

\author{
  \textbf{Nan Jia\textsuperscript{1}},
  \textbf{Haojin Yang\textsuperscript{2}},
  \textbf{Xing Ma\textsuperscript{3}},
  \textbf{Jiesong Lian\textsuperscript{1}},
  \textbf{Shuailiang Zhang\textsuperscript{3}}, \\
  \textbf{Weipeng Zhang\textsuperscript{3}},
  \textbf{Ke Zeng\textsuperscript{3}},
  \textbf{Xunliang Cai\textsuperscript{3}},
  \textbf{Zequn Sun\textsuperscript{4}}\thanks{Corresponding author.} \\
  \textsuperscript{1}Huazhong University of Science and Technology,
  \textsuperscript{2}Peking University, \\
  \textsuperscript{3}Meituan,
  \textsuperscript{4}Nanjing University \\
  {\small
    \texttt{jianan@hust.edu.cn},
    \texttt{yhj@stu.pku.cn},
    \texttt{lian700@hust.edu.cn},
    \texttt{zqsun.nju@gmail.com}
  } \\
  {\small
    \texttt{\{maxing08,zhangshuailiang,zhangweipeng02,zengke02,caixunliang\}@meituan.com}
  }
}

\begin{document}

\maketitle

\begin{figure*}[!t]
\centering
\begin{subfigure}[t]{0.34\textwidth}
    \centering
    \includegraphics[width=\linewidth]{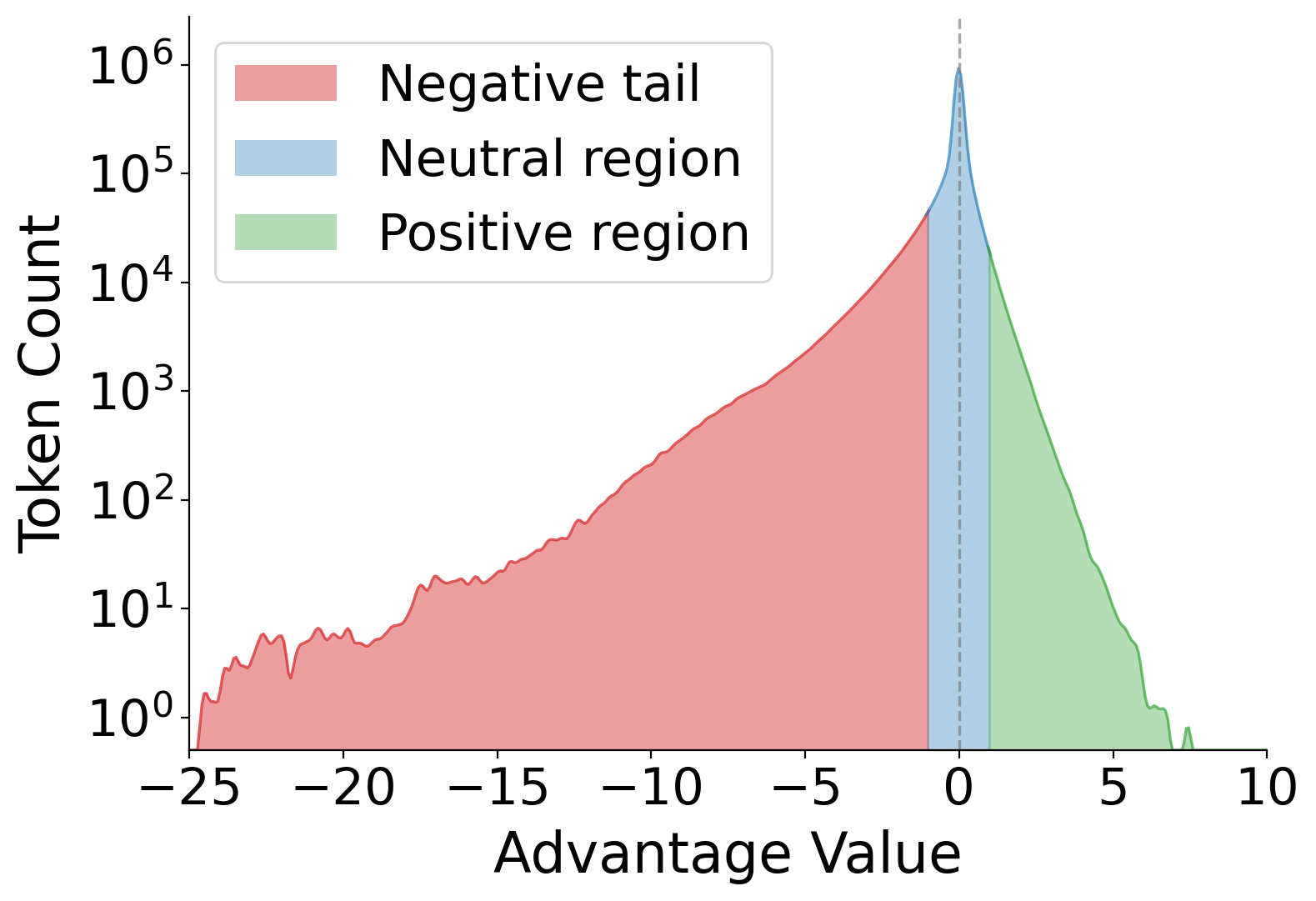}
    \caption{Advantage distribution.}
    \label{fig:kl_guidance_negative_neutral}
\end{subfigure}
\hfill
\begin{subfigure}[t]{0.32\textwidth}
    \centering
    \includegraphics[width=\linewidth]{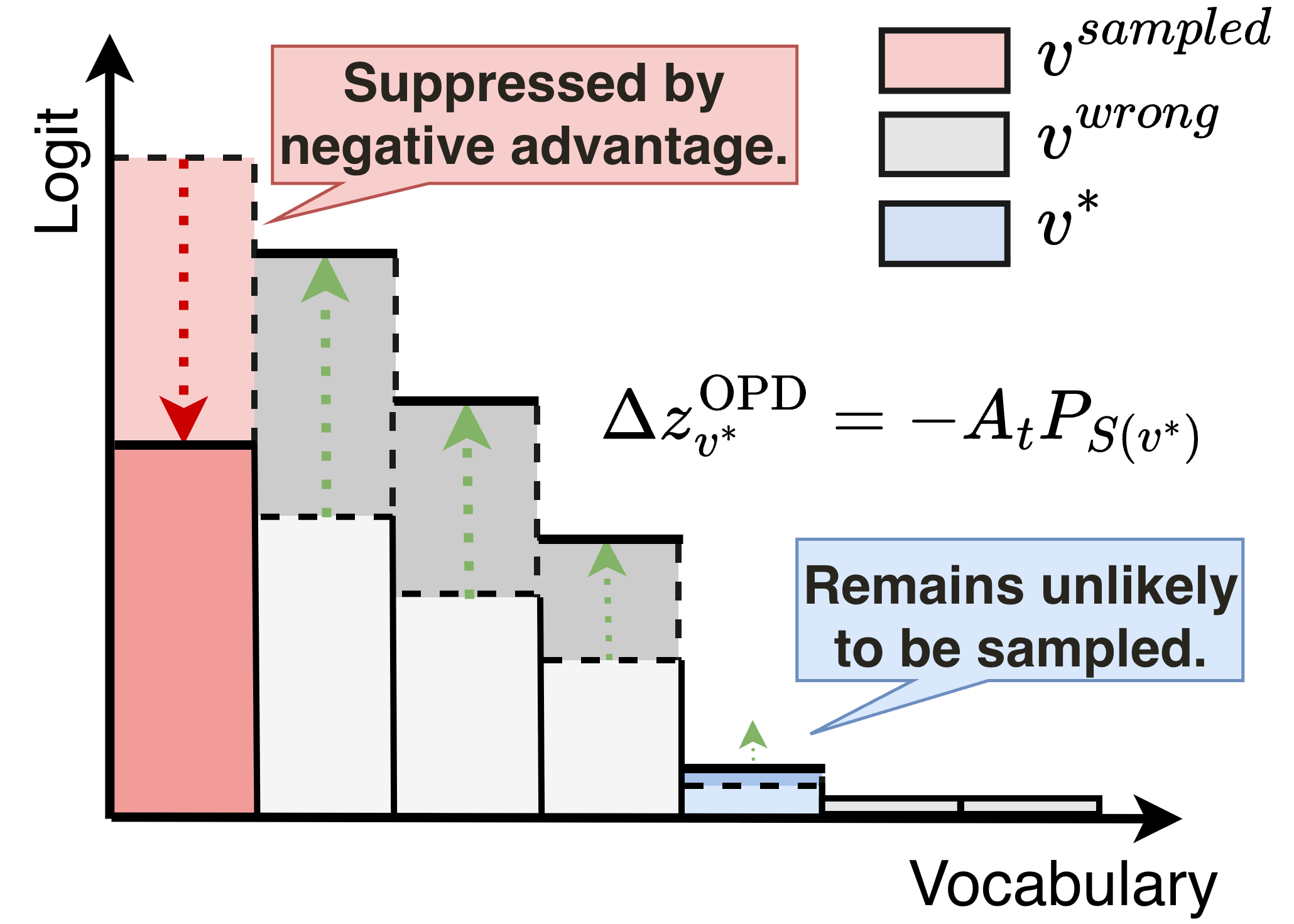}
    \caption{Exploration black hole.}
    \label{fig:exploration_black_hole}
\end{subfigure}
\hfill
\begin{subfigure}[t]{0.32\textwidth}
    \centering
    \includegraphics[width=\linewidth]{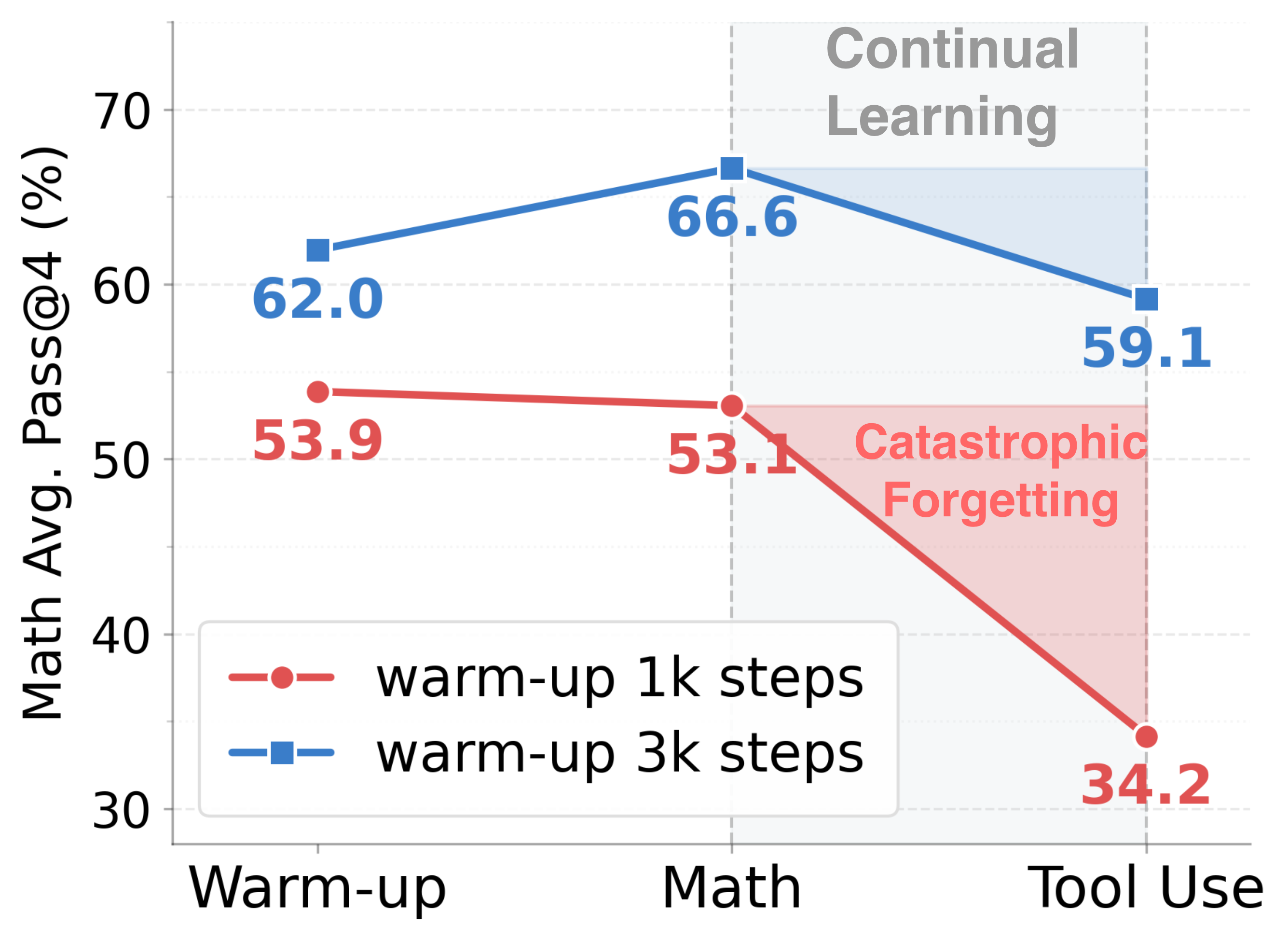}
    \caption{OPD under insufficient warm-up.}
    \label{fig:continual_forgetting}
\end{subfigure}
\caption{Observation and analysis of on-policy distillation.}
\label{fig:three_way_comparison}
\end{figure*}

\begin{abstract}
On-policy distillation (OPD) trains a student on its own trajectories with token-level teacher feedback and often outperforms off-policy distillation and reinforcement learning with verifiable rewards.
However, we find that its standard advantage-weighted policy gradient suffers from three structural weaknesses, including high-variance updates, vanishing gradients in zero-advantage regions, and exploration bottlenecks for models with limited capability.
We therefore propose asymmetric on-policy distillation (AOPD), which replaces ineffective policy gradient with localized divergence minimization in non-positive advantage regions while preserving positive reinforcement learning. 
Experiments on mathematical reasoning benchmarks show that AOPD consistently outperforms standard OPD, with average gains of 4.09 / 8.34 under strong / weak initialization, respectively. 
AOPD also maintains higher policy entropy during training and better capability retention during sequential tool-use adaptation.
Our source code is available at \href{https://github.com/stargazer319/AOPD}{repository}.
\end{abstract}

\section{Introduction}
\label{sec:introduction}

The strong capabilities of large language models (LLMs)~\citep{qwen3,kimi25,deepseekv4} come at a steep computational cost.
To make LLMs more accessible, knowledge distillation~\citep{hinton2015distilling} offers a principled path to compress the capabilities into smaller models.
However, this paradigm carries a well-known vulnerability in that the student encounters exposure bias~\citep{bengio2015scheduled} after training exclusively on the off-policy trajectories.

On-policy distillation (OPD)~\citep{thinkingmachines_opd} resolves this mismatch by optimizing the student on its own trajectories through a reinforcement learning objective. 
Using token-level rewards derived from teacher--student log-probability differences, it achieves strong training efficiency and has become a key technique in post-training~\citep{mimov2,glm5}.

However, we observe that this advantage formulation exhibits three limitations:
(1) In the negative region, the advantage distribution exhibits a heavy tail with large variance, as shown in Figure~\ref{fig:kl_guidance_negative_neutral}.
As a result, negative tokens can induce highly variable updates, thereby destabilizing the optimization process.
(2) A large fraction of sampled tokens concentrates near zero advantage.
The policy gradient consequently produces an almost vanishing update.
(3) OPD relies on the exploration mechanism of reinforcement learning for optimization. 
However, for models with limited initial capability, this process can easily degenerate into blind exploration.

To address the above limitations, we bridge the principles of reinforcement learning and supervised learning at the token level to propose asymmetric on-policy distillation (AOPD). 
Specifically, when a token receives a negative advantage, AOPD does not rely on self-exploration.
Instead, it directly matches the teacher model's target distribution conditioned on the on-policy trajectory prefix via Kullback-Leibler (KL) divergence.
This asymmetric design allows the student to retain the benefit of exploitation in reinforcement learning while receiving stronger correction at optimization bottlenecks.

We evaluate AOPD on competition-level mathematical reasoning benchmarks. 
Across model scales and warm-up settings, AOPD delivers stronger and more robust performance than existing baselines.
The localized guidance also leads to better capability retention in continual learning.
Our main contributions are summarized as follows:
\begin{itemize}
    \item We analyze on-policy distillation in zero and negative advantage regions and show that the standard advantage-based update provides limited learning signal for effective correction.
    \item We propose asymmetric on-policy distillation, a token-level method that adaptively shifts from policy gradient to localized distribution matching when the advantage signal carries limited information. 
    \item Experiments on mathematical reasoning benchmarks show that AOPD consistently improves over baselines, especially under weak initialization, and better preserves prior capabilities during sequential tool-use adaptation.
\end{itemize}


\section{Related Work}
\label{sec:related_work}
The foundation of knowledge distillation was established by \citet{hinton2015distilling}, who showed that a compact student can learn from a larger teacher by matching its softened output logits.
Classical off-policy distillation has achieved considerable success~\citep{distilling2022stepbystep,han2024tinyllm,xu2024surveykd} in strong-to-weak distillation.
For sequence generation, \citet{kim2016sequence} extended this idea to Sequence-Level Knowledge Distillation (SeqKD), aligning the student with the teacher's sequence-level distribution.
However, all these classical methods are trained off-policy on teacher-generated trajectories and therefore remain vulnerable to exposure bias~\citep{bengio2015scheduled}.

To mitigate this mismatch, recent work has shifted toward on-policy distillation, where the student learns on its own rollouts with teacher feedback. 
GKD~\citep{agarwal2024gkd} formalizes this as minimizing generalized $f$-divergences on student trajectories. 
\citet{gu2023minillm} and~\citet{thinkingmachines_opd} formulate on-policy distillation as a reinforcement learning objective with dense token-level rewards derived from the teacher.
\citet{li2026rethinkopd} identify teacher-student compatibility and novel capability transfer as two key conditions for effective OPD.
Beyond teacher alignment, ExOPD~\citep{yang2026learning} extrapolates reward scaling to surpass the teacher's capabilities.
SCOPE~\citep{zheng2026scope} further uses perplexity-calibrated adaptive weighting to apply different supervision to correct and incorrect trajectories.
Some work further explores multi-teacher extensions of OPD.
MiMo-V2~\citep{mimov2} introduces Multi-Teacher On-Policy Distillation (MOPD), while DeepSeek-V4~\citep{deepseekv4} adopts a similar pipeline via full-vocabulary reverse KL divergence.
CoPD~\citep{gu2026copd} further argues that post-hoc MOPD may fail to fully internalize teacher capabilities, and proposes bidirectional OPD to co-evolve experts.
Some work removes the reliance on external teachers by distilling from the model's own distributions.
OPSD~\citep{zhao2026opsd} and its long-context extension OPSDL~\citep{zhang2026opsdl} leverage ground-truth feedback on self-generated traces, while SDFT~\citep{shenfeld2025sdft} mitigates catastrophic forgetting under the same paradigm. 

\section{Preliminaries}
\label{sec:Preliminaries}
\subsection{Formulation of OPD}
OPD optimizes the reverse KL divergence with reinforcement learning based on the K1 estimator.
We formalize the training procedure below.
Given an input prompt $x$, let $y$ denote a response sampled from the student policy.
At step $t$, let $c_t \triangleq (x, y_{<t})$ denote the current context, and let $P_T(\cdot \mid c_t)$ and $P_S(\cdot \mid c_t)$ denote the teacher and student conditional distributions.
The advantage of token $y_t \sim P_S(\cdot \mid c_t)$ is defined as:
\begin{equation}
A_t = \operatorname{sg}\!\left[\log P_T(y_t \mid c_t) - \log P_S(y_t \mid c_t)\right],
\label{eq:advantage}
\end{equation}
where $\operatorname{sg}[\cdot]$ denotes the stop-gradient operator.
The corresponding training loss is
\begin{equation}
\mathcal{L}_{\mathrm{OPD}}
=
-\mathbb{E}\!\left[
\frac{1}{|y|}\sum_{t=1}^{|y|}
A_t \log P_S(y_t \mid c_t)
\right].
\label{eq:opd}
\end{equation}

Motivated by \citet{zhu2025surprising} and \citet{tang2025samplepolarity}, which highlight the distinct roles of positive and negative samples in reinforcement learning, we decompose OPD into positive and negative reinforcement at the token level.
Let $S^{+}$ and $S^{-}$ denote the sets of token positions with positive and negative advantages, respectively, and let $y_t^{+}$ and $y_t^{-}$ denote the corresponding sampled tokens.
The OPD loss can be written as
\begin{align}
\mathcal{L}_{\mathrm{OPD}}
&=
\underbrace{
-\mathbb{E}\!\left[
\frac{1}{|y|}
\sum_{t \in S^{+}}
A_t \log P_S(y_t^{+} \mid c_t)
\right]
}_{\mathcal{L}_{\mathrm{Pos}}}
\nonumber \\
&
\underbrace{
-\mathbb{E}\!\left[
\frac{1}{|y|}
\sum_{t \in S^{-}}
A_t \log P_S(y_t^{-} \mid c_t)
\right]
}_{\mathcal{L}_{\mathrm{Neg}}}.
\label{eq:opd_decompose}
\end{align}
Here, $\mathcal{L}_{\mathrm{Pos}}$ corresponds to positive reinforcement, which reinforces teacher-favored sampled tokens.
In contrast, $\mathcal{L}_{\mathrm{Neg}}$ corresponds to negative reinforcement, which suppresses teacher-disfavored sampled tokens and promotes exploration.

\subsection{Limitations of OPD}
\label{sec:insight}

Despite its intuitive formulation, standard OPD remains subject to several challenges in practice.
Through theoretical analysis and experimental validation, we identify three characteristic limitations.

\paragraph{Heavy Tails in Negative Advantages.}
Figure~\ref{fig:kl_guidance_negative_neutral} shows that the negative side of the advantage distribution exhibits a substantially broader tail.
The extreme variance originates from the logarithmic nature of the advantage formulation.
The difference between the log probability of the teacher and the student amplifies exponentially as the probability approaches zero.
Because the update magnitude is directly scaled by this unbounded scalar, the mechanism artificially inflates gradient variance and renders the overall optimization fragile.

\paragraph{Vanishing Updates at Zero Advantages.}
A second limitation arises in the neutral regime.
As shown in Figure~\ref{fig:kl_guidance_negative_neutral}, the vast majority of sampled tokens cluster around an advantage of zero. 
In this regime, the teacher remains able to offer fine-grained preferences over alternative reasoning paths.
However, OPD only receives a zero signal, producing an almost vanishing update.

\paragraph{Exploration Black Hole.}
A fundamental optimization problem arises in negative reinforcement, where the update of OPD is restricted in both direction and magnitude.
To understand this behavior, we examine the OPD update at the logit level.
Let $z_v$ denote the student logit for token $v$ under context $c_t$, so that
$P_S(v \mid c_t) = \mathrm{softmax}(z)_v$.
By the softmax Jacobian, the update for any token $v$ is given by:
\begin{equation}
\Delta z_v^{\mathrm{OPD}} \propto A_t \bigl(\mathbb{I}(v = y_t) - P_S(v \mid c_t)\bigr).
\label{eq:opd_update}
\end{equation}
When $A_t < 0$, the sampled token is penalized, and for every unsampled token $v \neq y_t$, the update becomes
\begin{equation}
\Delta z_v^{\mathrm{OPD}} \propto -A_t P_S(v \mid c_t) > 0.
\label{eq:opd_unsampled}
\end{equation}
Hence, the released probability mass is redistributed strictly according to the current prior of the student.
As illustrated in Figure~\ref{fig:exploration_black_hole}, for an essential token $v^*$ with negligible prior probability $P_S(v^* \mid c_t) \approx 0$, the corrective update is correspondingly negligible.
OPD suppresses the sampled mistake without effectively promoting the correct alternative, trapping the model in an exploration black hole.
This pathology is more severe for models with weaker foundational capabilities, as essential tokens are more likely to reside in the low-probability tail of the student prior. 
Figure~\ref{fig:continual_forgetting} illustrates this effect by comparing OPD models with different steps of SFT warm-up.
The model with fewer warm-up steps makes exploration black holes more likely and thereby undermines both reasoning improvement and ability retention.

These observations suggest that tokens in non-positive regions should not be optimized in the same way as positive reinforcement.
The learning signal in these regions is insufficient and therefore calls for an asymmetric training rule.

\begin{figure*}[t]
    \centering
    \includegraphics[width=\textwidth]{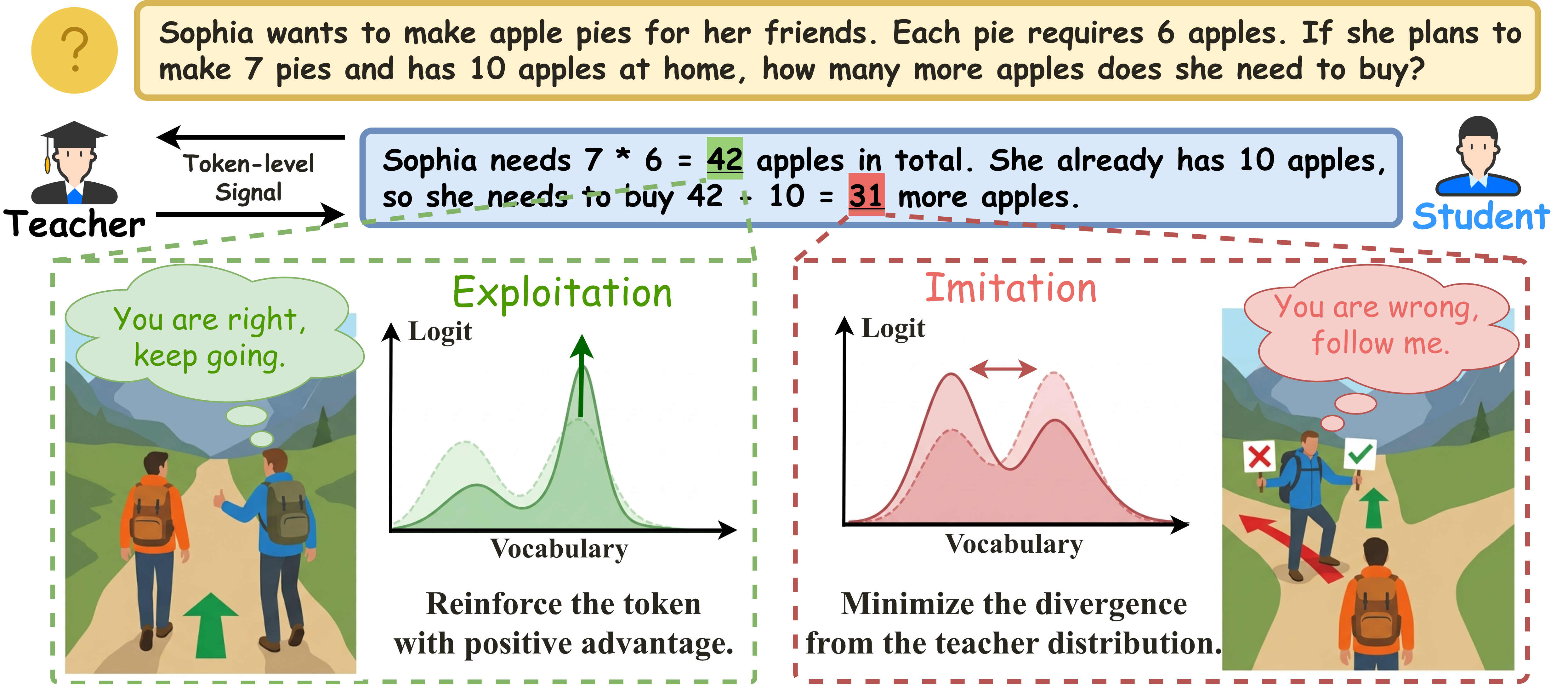}
    \caption{Overview of AOPD.
    Left (exploitation): When the student's reasoning aligns with the teacher, AOPD reinforces successful exploration.
    Right (imitation): When the student encounters an exploration bottleneck, AOPD actively switches to directed teacher guidance by minimizing the distributional divergence.}
    \label{fig:overview}
\end{figure*}
\section{Asymmetric On-Policy Distillation}
\label{sec:method}

To systematically resolve the structural vulnerability of standard on-policy distillation, we propose asymmetric on-policy distillation (AOPD). 
As illustrated in Figure~\ref{fig:overview}, AOPD dynamically modulates the learning paradigm at the token level.
It shifts between reinforcement learning for exploitation and soft-label supervised learning for imitation based on the student's local capability.

\subsection{The AOPD Framework}
\label{subsec:framework}
We now present a formal definition of AOPD.
We compute the token-level advantage $A_t$ using Eq.~\ref{eq:advantage} as in standard OPD. 
As discussed in Section~\ref{sec:insight}, the main optimization difficulties of OPD are concentrated in negative and zero-advantage regions.
We therefore replace the update in non-positive regions with forward KL minimization computed on the teacher-defined support.
For each position $t$, we define the teacher support as
\begin{equation}
S_t = \mathrm{TopK}\!\left(P_T(\cdot \mid c_t), K\right),
\label{eq:topk_support}
\end{equation}
where $S_t$ contains the $K$ highest-probability tokens under the teacher distribution.
On this support, the forward KL guidance loss is instantiated as
\begin{align}
\mathcal{L}^{\mathrm{FKL}}_t
=&
\frac{1}{K}
\sum_{v \in S_t}
P_T(v \mid c_t)
\Big(
\log P_T(v \mid c_t) \nonumber \\
& - 
\log P_S(v \mid c_t)
\Big).
\label{eq:aopd_fkl}
\end{align}
Accordingly, the AOPD objective is defined as
\begin{align}
\mathcal{L}_{\mathrm{AOPD}}
= &
\mathbb{E}\!\left[
\frac{1}{|y|}
\sum_{t \in S^{\le 0}}
\mathcal{L}^{\mathrm{FKL}}_t
\right]
+
\mathcal{L}_{\mathrm{Pos}},
\label{eq:aopd_core}
\end{align}
where $S^{\le 0}$ denotes the non-positive token set.

AOPD uses zero advantage as the default intervention boundary.
To define the intervention location in a more general form, we further introduce a threshold parameter $\tau$.
Considering the logarithmic nature of the standard advantage formulation and its unbounded variance, we use the bounded probability difference to determine whether to trigger teacher intervention.
Specifically, we define the token-level mask as:
\begin{equation}
G_t = \mathbb{I}\!\left(P_T(y_t \mid c_t) - P_S(y_t \mid c_t) \le \tau \right),
\label{eq:indicator}
\end{equation}
where $\mathbb{I}(\cdot)$ denotes the indicator function.
The generalized AOPD objective is then defined as:
\begin{align}
\mathcal{L}_{\mathrm{AOPD}}
=&
\mathbb{E}\biggl[
\frac{1}{|y|}\sum_{t=1}^{|y|}
\Bigl(
G_t \mathcal{L}^{\mathrm{FKL}}_t \nonumber \\
& + (1-G_t)\mathcal{L}^{\mathrm{OPD}}_t
\Bigr)
\biggr].
\label{eq:AOPD_total}
\end{align}
Setting \(\tau=-1\) disables intervention for all tokens and reduces AOPD to OPD, while setting \(\tau=1\) applies global supervised distribution matching and recovers a GKD objective.
In our method, we set $\tau = 0$, so that intervention is applied exactly to the non-positive regime.
The effects of alternative intervention locations and different values of $\tau$ are further analyzed in Section~\ref{subsec:kl_guidance} and Appendix~\ref{sec:ablation_tau}.


\begin{figure*}[ht]
\centering
\begin{subfigure}[t]{0.32\textwidth}
    \centering
    \includegraphics[width=\linewidth]{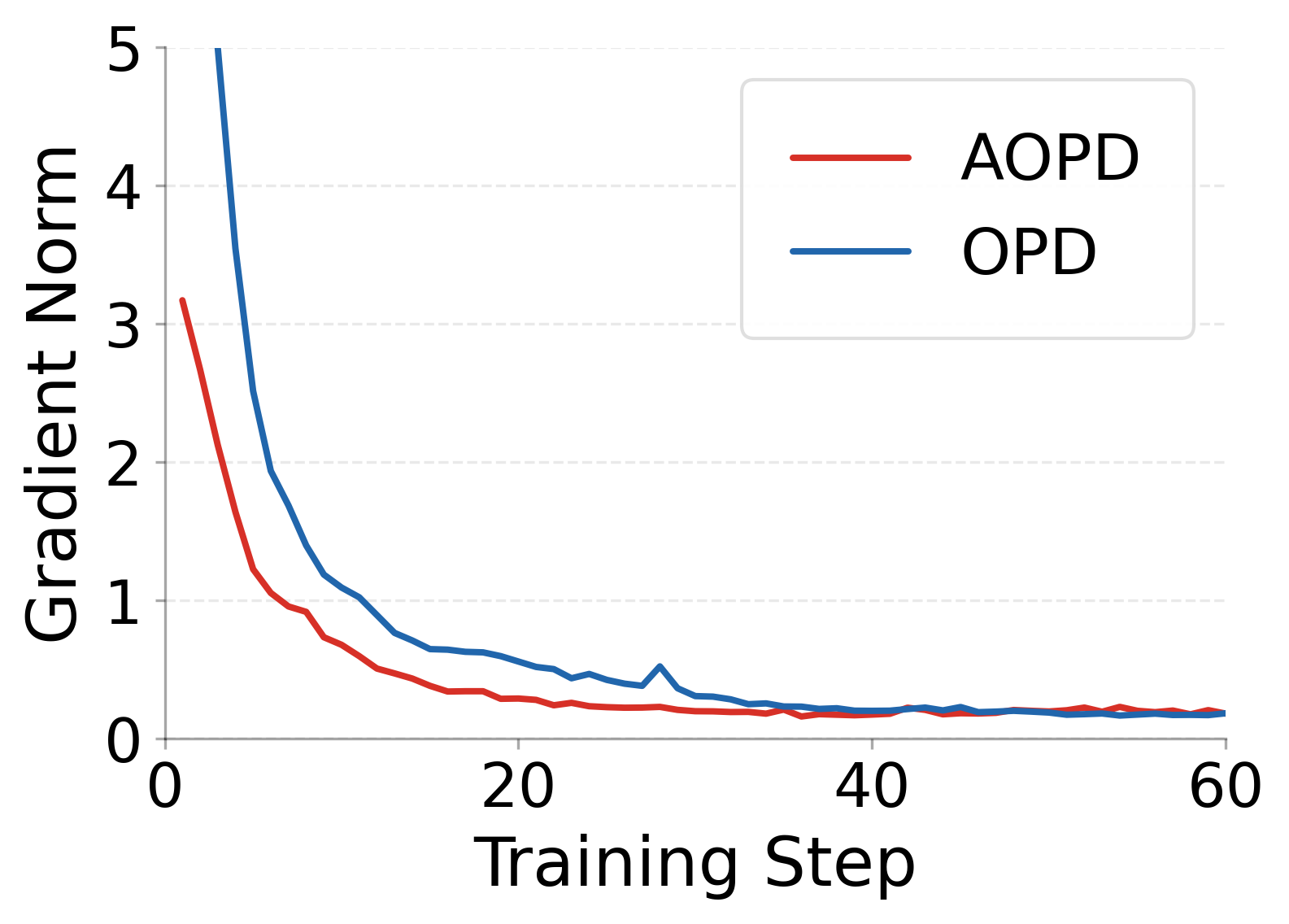}
    \caption{Gradient norm during training.}
    \label{fig:aopd_joint_gradnorm}
\end{subfigure}
\hfill
\begin{subfigure}[t]{0.32\textwidth}
    \centering
    \includegraphics[width=\linewidth]{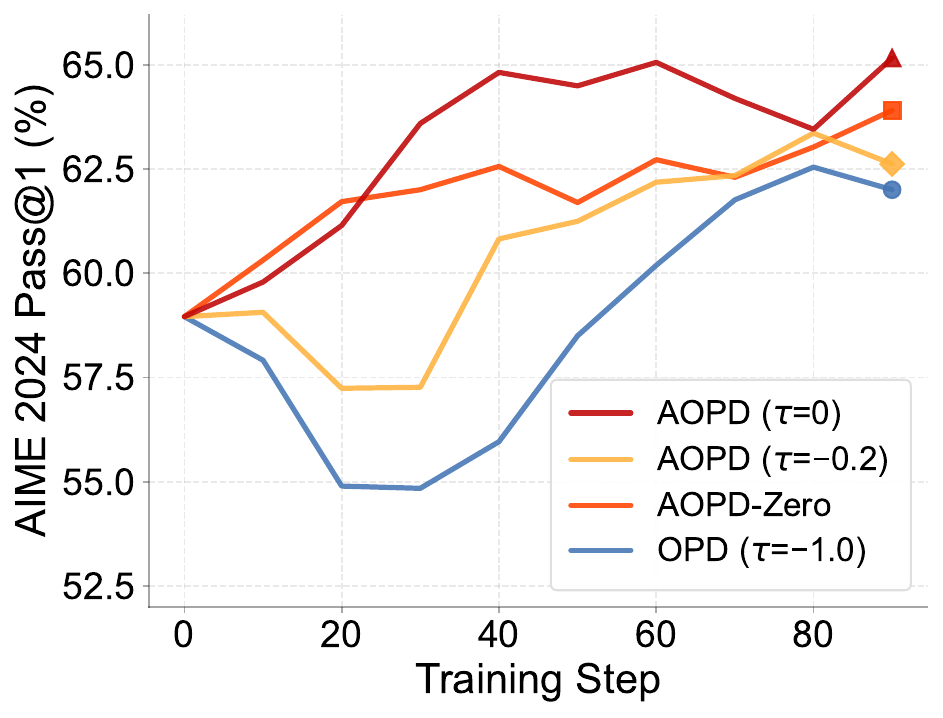}
    \caption{AIME 2024 Pass@1 training curves under different intervention strategies.}
    \label{fig:aopd_joint_zero}
\end{subfigure}
\hfill
\begin{subfigure}[t]{0.32\textwidth}
    \centering
    \includegraphics[width=\linewidth]{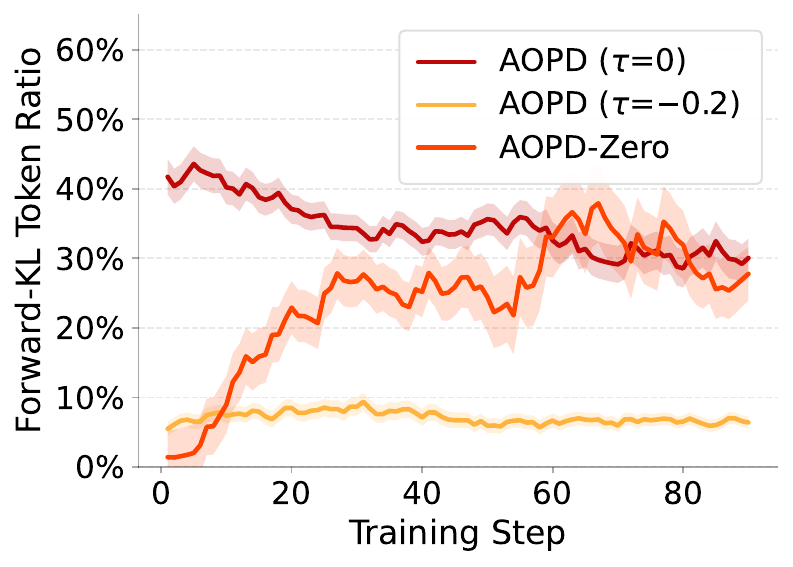}
    \caption{Ratio of tokens receiving forward KL guidance during training.}
    \label{fig:aopd_joint_ratio}
\end{subfigure}
\caption{Training dynamics under different divergence-guidance strategies.}
\label{fig:aopd_joint_validation}
\end{figure*}

\subsection{Truncated Forward-KL Guidance}
\label{subsec:topk_and_divergence}
In principle, the intervention objective should minimize the divergence between the teacher and the student over the full vocabulary.
The divergence family can be written in the general form of Jensen--Shannon divergence (JSD) as
\begin{align}
\mathrm{JSD}_{\beta}(P_T \parallel P_S)
=&
\beta D_{\mathrm{KL}}(P_T \parallel P_M) \nonumber \\
&+
(1-\beta)D_{\mathrm{KL}}(P_S \parallel P_M),
\label{eq:jsd}
\end{align}
where $P_M = \beta P_T + (1-\beta)P_S$ and $\beta \in [0,1]$.
Here, $\beta=1$ corresponds to forward KL, while $\beta=0$ corresponds to reverse KL.

However, evaluating such a divergence at every intervened position is computationally prohibitive in large-vocabulary LLM training.
We therefore instantiate the guidance term on the teacher-selected top-$K$ support in Eq.~\ref{eq:topk_support}.
Under top-$K$ truncation, the learning objective becomes a correction defined on a teacher-selected domain.
In this setting, the teacher specifies both the candidate tokens and the reference distribution within that domain.
Therefore, forward KL is the natural choice because it preserves the same teacher-conditioned measure in both support construction and loss weighting.
In contrast, reverse KL would reweight the update according to the student's current distribution on a domain selected by the teacher.
The detailed gradient analysis is provided in Appendix~\ref{sec:appendix_fkl_gradient}.

Empirical evidence in Section~\ref{sec:ablation_beta} further favors placing more emphasis on the teacher distribution.
We also find that reverse KL induces anomalous training behaviors in Appendix~\ref{sec:beta_training_dynamics}.
Combined with the support-consistency argument above, these results motivate our choice of $\beta=1$.


\begin{table*}[t]
\centering
\small
\begin{tabular}{lccccccccc}
\toprule
\multicolumn{1}{c}{\multirow{2}{*}{\textbf{Method}}} & \multicolumn{2}{c}{\textbf{AIME 2024}} & \multicolumn{2}{c}{\textbf{AIME 2025}} & \multicolumn{2}{c}{\textbf{HMMT 2025 (Feb)}} & \multicolumn{2}{c}{\textbf{Average}} \\
\cmidrule(lr){2-3} \cmidrule(lr){4-5} \cmidrule(lr){6-7} \cmidrule(lr){8-9}
& \textit{Pass@1} & \textit{Pass@4} & \textit{Pass@1} & \textit{Pass@4} & \textit{Pass@1} & \textit{Pass@4} & \textit{Pass@1} & \textit{Pass@4} \\

\midrule
\multicolumn{9}{l}{\textbf{Qwen3-4B-Base}} \\
\quad \textit{-- 3k SFT Warm-up} \\
\quad \quad SFT Checkpoint   & 52.29 & 71.90 & 46.25 & 64.82 & 27.08 & 38.97 & 41.87 & 58.56 \\
\quad \quad + SeqKD          & 42.50 & 65.23 & 40.00 & 60.09 & 24.58 & 38.01 & 35.69 & 54.44 \\
\quad \quad + OPD            & 57.08 & \textbf{76.69} & \underline{51.88} & 68.12 & \textbf{32.92} & \textbf{47.65} & 47.29 & \textbf{64.15} \\
\quad \quad + ExOPD          & \underline{59.17} & 73.99 & 51.25 & 66.59 & \underline{32.08} & 47.30 & \underline{47.50} & 62.63 \\
\quad \quad + GKD            & 58.13 & \underline{76.61} & 50.42 & \underline{69.12} & 29.79 & 45.39 & 46.11 & 63.71 \\
\quad \quad + \textbf{AOPD}  & \textbf{61.04} & 75.08 & \textbf{54.37} & \textbf{69.73} & \underline{32.08} & \underline{47.40} & \textbf{49.16} & \underline{64.07} \\

\midrule
\multicolumn{9}{l}{\textbf{Qwen3-8B-Base}} \\
\quad \textit{-- 1k SFT Warm-up} \\
\quad \quad SFT Checkpoint   & 45.21 & 68.29 & 39.79 & 58.62 & 24.58 & 34.68 & 36.53 & 53.86 \\
\quad \quad + SeqKD          & 40.21 & 68.08 & 30.63 & 50.00 & 20.62 & 36.78 & 26.11 & 42.87 \\
\quad \quad + OPD            & 43.13 & 67.38 & 38.12 & 56.84 & 23.13 & 34.95 & 34.79 & 53.06 \\
\quad \quad + ExOPD          & 45.83 & 72.31 & 36.04 & 53.86 & 24.37 & 38.02 & 35.41 & 54.73 \\
\quad \quad + GKD            & \textbf{54.79} & \textbf{77.92} & \underline{42.08} & \underline{63.69} & \underline{25.83} & \underline{40.21} & \underline{40.90} & \underline{60.61} \\
\quad \quad + \textbf{AOPD}  & \underline{53.75} & \underline{77.39} & \textbf{45.21} & \textbf{66.86} & \textbf{30.42} & \textbf{46.73} & \textbf{43.13} & \textbf{63.66} \\
\addlinespace
\quad \textit{-- 3k SFT Warm-up} \\
\quad \quad SFT Checkpoint   & 58.96 & 74.95 & 47.71 & 66.73 & 31.04 & 44.19 & 45.90 & 61.96 \\
\quad \quad + SeqKD          & 51.04 & 76.56 & 40.00 & 56.98 & 25.42 & 43.80 & 38.82 & 59.11 \\
\quad \quad + OPD            & 61.46 & 78.61 & \underline{52.29} & \underline{70.44} & \underline{34.17} & \underline{50.86} & 49.31 & 66.64 \\
\quad \quad + ExOPD          & 61.67 & 77.55 & 48.75 & 65.83 & \underline{34.17} & 50.22 & 48.20 & 64.53 \\
\quad \quad + GKD            & \underline{66.25} & \textbf{80.65} & 48.54 & \textbf{70.56} & \underline{34.17} & 49.67 & \underline{49.65} & \underline{66.96} \\
\quad \quad + \textbf{AOPD}  & \textbf{66.87} & \underline{79.58} & \textbf{55.00} & 70.28 & \textbf{38.33} & \textbf{52.42} & \textbf{53.40} & \textbf{67.43} \\
\bottomrule
\end{tabular}
\caption{
  Main results across multiple math reasoning benchmarks. Baseline capabilities before on-policy distillation are given by the SFT Checkpoint. We compare AOPD against SeqKD, OPD, GKD and ExOPD. \textbf{Bold} indicates the best result within each configuration block, while \underline{underlined} values denote the second best.
}
\label{tab:main_math}
\end{table*}

\subsection{A Unified Solution to the Bottlenecks}
\label{subsec:kl_guidance}

We now explain why the single confidence-based rule in AOPD is sufficient to unify the solutions for the three optimization bottlenecks identified in Section~\ref{sec:insight}. 
By setting $\tau=0$, the core mechanism is to switch from policy gradient to direct distribution matching exactly where $A_t \le 0$.

\paragraph{(1) Eliminating high-variance negative advantages.}
As analyzed in Section~\ref{sec:insight}, standard OPD suffers from high-variance gradient updates when $A_t < 0$.
AOPD eliminates this instability by halting the policy gradient update in these regions and replacing it with divergence guidance.
For the forward KL case, the logit-level correction is instead governed by the bounded distribution gap:
\begin{align}
\Delta z_v^{\mathrm{AOPD}} \propto P_T(v \mid c_t) - P_S(v \mid c_t).
\label{eq:aopd_distribution_gap}
\end{align}
Constrained by the probability simplex, this signal naturally caps the maximum update magnitude. 
Figure~\ref{fig:aopd_joint_gradnorm} confirms this effect, showing that AOPD maintains a substantially smaller gradient norm than OPD in early stage of training.

\paragraph{(2) Recovering gradients in zero-advantage regions.}
In zero-advantage regions where the original gradient nearly vanishes, AOPD still aligns the student toward alternative reasoning paths.
Consequently, the student can still receive meaningful corrections on the teacher-defined support.
Figure~\ref{fig:aopd_joint_zero} directly supports this by evaluating AOPD-Zero, a variant that applies KL-divergence intervention only to tokens whose advantages are zero.
Compared with standard OPD, this targeted replacement already yields clear gains, confirming that neutral regions contain useful teacher signals that are discarded by the policy gradient method.

\paragraph{(3) Escaping exploration bottlenecks through teacher guidance.}
When trapped in an exploration black hole, OPD merely suppresses the sampled mistake without promoting the correct, unsampled token.
AOPD resolves this by explicitly injecting the missing directional signal. 
As shown in Eq.~\ref{eq:aopd_distribution_gap}, any token favored by the teacher but underestimated by the student immediately receives a positive correction proportional to their probability gap.
By directly matching the teacher's target distribution on the on-policy prefix, AOPD repairs optimization bottlenecks instantly rather than waiting for the student to discover the correct token.
This is further evidenced in Figure~\ref{fig:aopd_joint_zero} by the $\tau=-0.2$ variant, which applies forward KL guidance only when the teacher--student gap is sufficiently negative.
Figure~\ref{fig:aopd_joint_ratio} shows that this variant intervenes on fewer than $10\%$ of all generated tokens.
However, this sparse intervention already yields clear gains over OPD, confirming that targeted teacher distribution guidance is highly efficient at rescuing optimization from exploration black holes.



\section{Experiments}

\subsection{Experimental Setup}
\label{sec:exp_setup}

\paragraph{Training Data and Benchmarks.}
Our experiments focus on two abilities: mathematical reasoning and tool use. 
For mathematical reasoning, we first use the OpenThoughts dataset~\citep{guha2025openthoughts} for warm-up training, and then conduct distillation on DeepMath~\citep{deepmath2025}. 
We evaluate mathematical reasoning on AIME 2024/2025 and HMMT 2025 (Feb). 
For tool use, we further train the model on ToolAlpaca~\citep{tang2023toolalpaca} in the continual learning stage, and evaluate it on the test set with the tool-call success rate. 
This setting allows us to examine both reasoning distillation performance and the ability to acquire new skills while retaining previously learned reasoning capability.

\paragraph{Models.}
Following convention, we evaluate our method under two settings: (1) distilling from Qwen3-32B to Qwen3-8B-Base, and (2) distilling from Qwen3-8B to Qwen3-4B-Base~\citep{qwen3}. 
These challenging scenarios validate our method across different parameter scales.

\paragraph{Baselines.}
We compare AOPD against four baselines:
(i) \textbf{SeqKD}~\citep{kim2016sequence}, which continues supervised fine-tuning on the trajectories generated by the teacher model;
(ii) \textbf{OPD}~\citep{thinkingmachines_opd}, standard on-policy distillation using K1-estimated reverse KL divergence as the per-token advantage;
(iii) \textbf{GKD}~\citep{agarwal2024gkd}, which we implement as forward KL divergence on the teacher's top-$K$ support over student-generated trajectories; and (iv) \textbf{ExOPD}~\citep{yang2026learning}, an extrapolation-based variant for OPD.

\paragraph{Training Pipeline.}
We first warm up the base models to equip them with basic abilities.
Specifically, we warm up base models for 3,000 steps, alongside an experimental setting of 1,000 steps to yield variants with distinct foundational capabilities for robustness assessment.
At distillation stage, models are first trained on mathematical tasks, followed by tool-call tasks for continual learning.
During the subsequent distillation phase, we uniformly set the teacher support size $K$ to 32 in AOPD and GKD for a fair comparison.
Regarding ExOPD, we use its best-reported hyperparameters in the paper.
More details about implementation and training hyperparameters are reported in Appendix~\ref{sec:implementation_detail}.

\subsection{Main Results on Reasoning}
\label{subsec:main_math}

Table~\ref{tab:main_math} shows that AOPD consistently achieves the best overall reasoning performance across model scales and warm-up settings. 
For Qwen3-4B-Base, it already attains the best average Pass@1, demonstrating its effectiveness in smaller-scale distillation. 
For Qwen3-8B-Base with the more challenging 1k-step warm-up, AOPD still delivers the best overall performance, while standard OPD falls below the warm-up baseline, highlighting AOPD's robustness. 
Under the 3k-step warm-up, AOPD further achieves the best overall results, reaching 53.40 average Pass@1 and 67.43 average Pass@4.
Extended results are provided in Appendix~\ref{sec:extended_main_result}.

\begin{figure}[h]
    \centering
    \begin{subfigure}[t]{0.48\columnwidth}
        \centering
        \includegraphics[width=\linewidth]{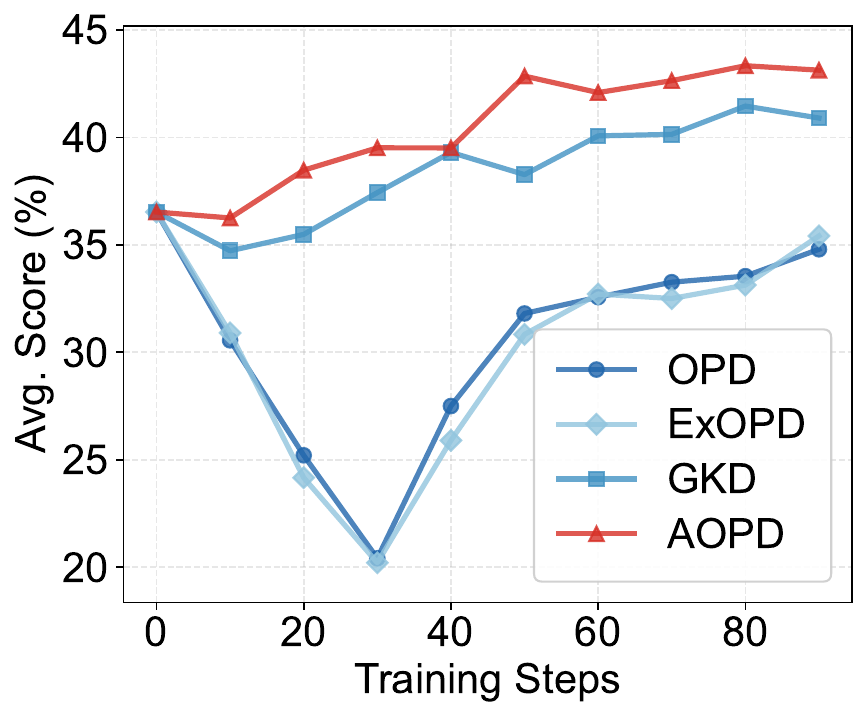}
        \caption{Weak initialization.}
        \label{fig:dynamics_base}
    \end{subfigure}
    \hfill
    \begin{subfigure}[t]{0.48\columnwidth}
        \centering
        \includegraphics[width=\linewidth]{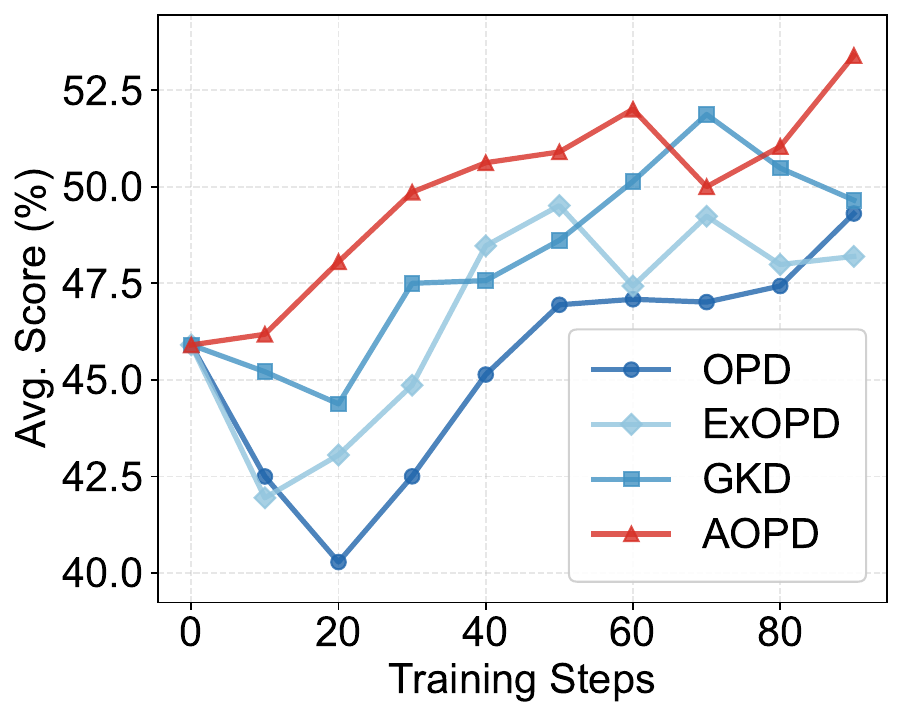}
        \caption{Strong initialization.}
        \label{fig:dynamics_advanced}
    \end{subfigure}
    \caption{Training dynamics of Qwen3-8B-Base under different initializations.}
    \label{fig:training_dynamics}
\end{figure}

Figure~\ref{fig:training_dynamics} further illustrates the training dynamics of different methods in terms of average math score. 
Across both settings, OPD and ExOPD exhibit a noticeable early performance drop. 
This issue is especially severe under the weaker 1k-step initialization, where recovery from poor states is difficult.
GKD shows a more stable trajectory, but its final performance still falls short of AOPD. 
In contrast, AOPD improves steadily from the start under both initialization settings and consistently reaches the highest performance ceiling.
Its asymmetric design not only avoids the early degradation but also translates training stability into stronger final reasoning performance.
Detailed training dynamics are further shown in Appendix~\ref{sec:detailed_training}.

\subsection{Continual Learning}
\label{subsec:cl_tools}
We evaluate continual learning on ToolAlpaca to test whether models can acquire a new ability while retaining mathematical reasoning performance. 
Table~\ref{tab:cl_results} shows that AOPD achieves the best balance between adaptation and retention, attaining a competitive 75.0\% success rate while preserving the strongest math performance. 
In contrast, OPD, ExOPD, and GKD all suffer math degradation after continual learning, with OPD showing the largest drop. 
These results suggest that localized teacher guidance enables adaptation to new abilities without overwriting prior reasoning ability.

\begin{table}[ht]
\centering
\resizebox{\columnwidth}{!}{
\begin{tabular}{lccc}
\toprule
\textbf{Method} & \textbf{Tool Use} & \begin{tabular}[c]{@{}c@{}}\textbf{Math Avg. Pass@4} \\ \textbf{Before / After}\end{tabular} & \textbf{Drop ($\downarrow$)} \\
\midrule
OPD   & 65.0 & 66.64 / 59.15 & -7.49 \\
ExOPD & 67.0 & 64.53 / 63.04 & -1.49 \\
GKD   & \textbf{76.0} & 66.96 / 64.54 & -2.42 \\
\textbf{AOPD} & 75.0 & \textbf{67.43 / 68.04} & \textbf{+0.61} \\
\bottomrule
\end{tabular}
}
\caption{Performance after continual learning.}
\label{tab:cl_results}
\end{table}

\begin{figure}[t]
\centering
\begin{minipage}[t]{0.48\columnwidth}
    \centering
    \includegraphics[width=\linewidth]{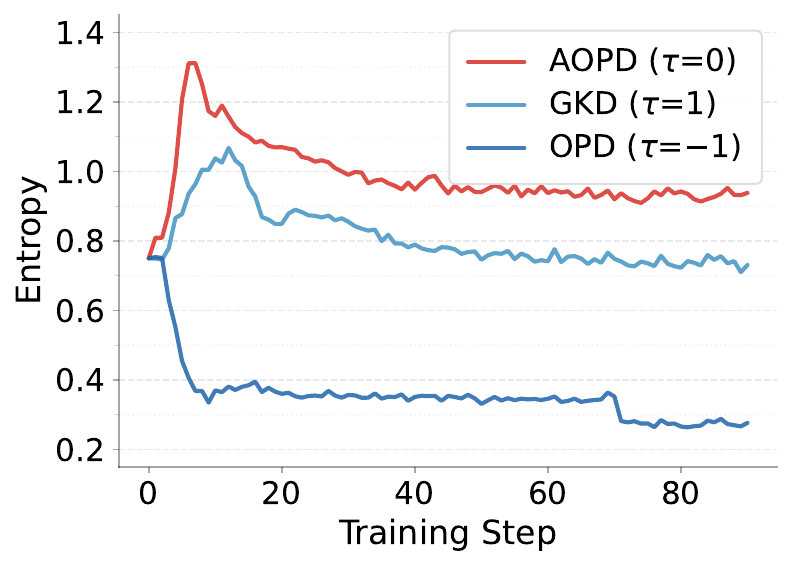}
    \captionof{figure}{Policy entropy during training.}
    \label{fig:entropy_comparison}
\end{minipage}\hfill
\begin{minipage}[t]{0.48\columnwidth}
    \centering
    \includegraphics[width=\linewidth]{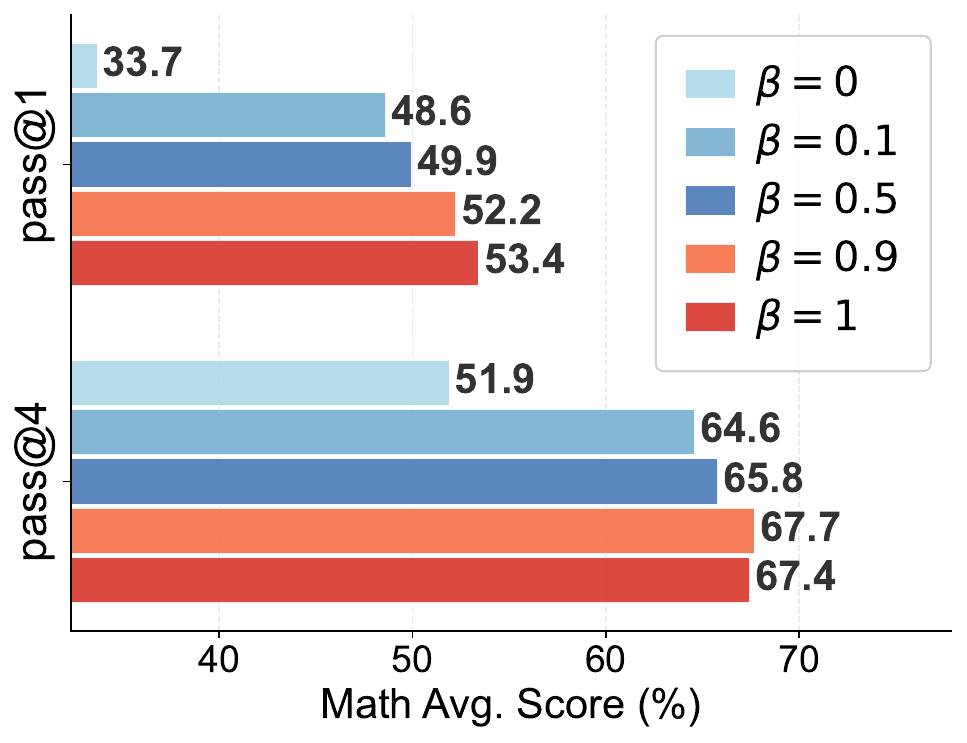}
    \captionof{figure}{Effect of the divergence weighting.}
    \label{fig:ablation_beta}
\end{minipage}
\end{figure}

We further examine the training dynamics of AOPD to explain its capability retention.
As earlier shown in Figure~\ref{fig:aopd_joint_ratio}, the fraction of tokens receiving divergence guidance is limited and gradually decreases from about 40\% to 30\% during training.
Most tokens are still optimized by policy gradient. Figure~\ref{fig:entropy_comparison} further shows that OPD suffers severe entropy collapse, while GKD keeps the policy near its initial entropy. 
In contrast, AOPD maintains higher entropy throughout training, suggesting that it preserves a broader policy space while applying correction only at difficult positions. 
This helps explain why AOPD better acquires new tool-use ability while retaining prior reasoning competence.
In other words, AOPD does not rely on globally reshaping the student into the teacher, but on selectively repairing the difficult parts of the trajectory where explicit guidance is most needed.

\begin{figure*}[t]
    \centering
    \begin{subfigure}[t]{0.32\textwidth}
        \centering
        \includegraphics[width=\linewidth]{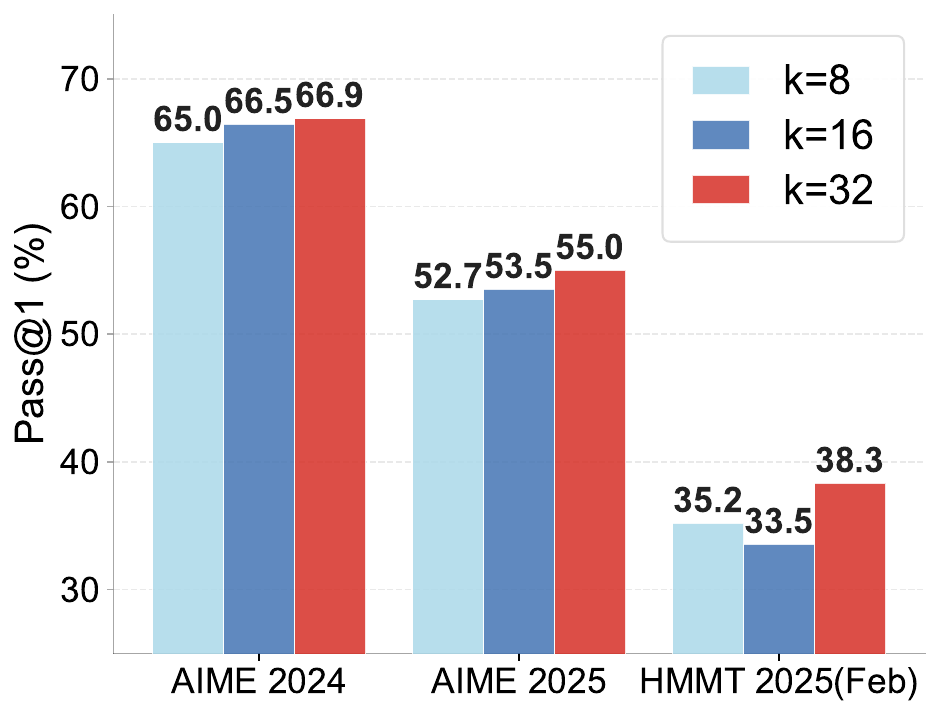}
        \caption{Scores of different top-$K$.}
        \label{fig:ablation_topk_bar}
    \end{subfigure}
    \hfill
    \begin{subfigure}[t]{0.32\textwidth}
        \centering
        \includegraphics[width=\linewidth]{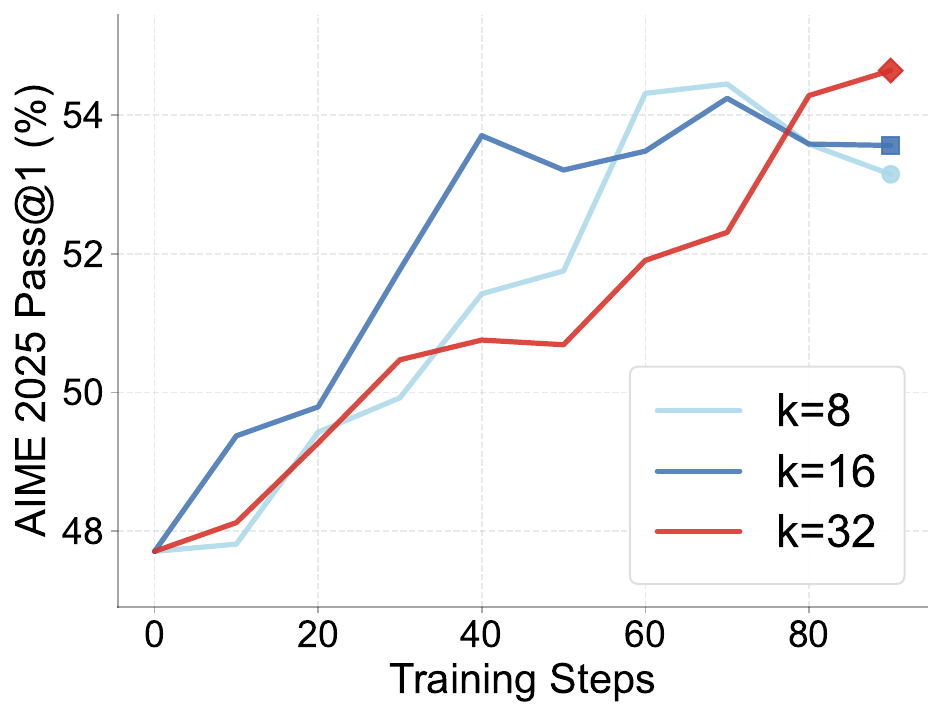}
        \caption{Training dynamics of different top-$K$.}
        \label{fig:ablation_topk_aime25}
    \end{subfigure}
    \label{fig:k_ablation}
    \hfill
    \begin{subfigure}[t]{0.32\textwidth}
        \centering
        \includegraphics[width=\linewidth]{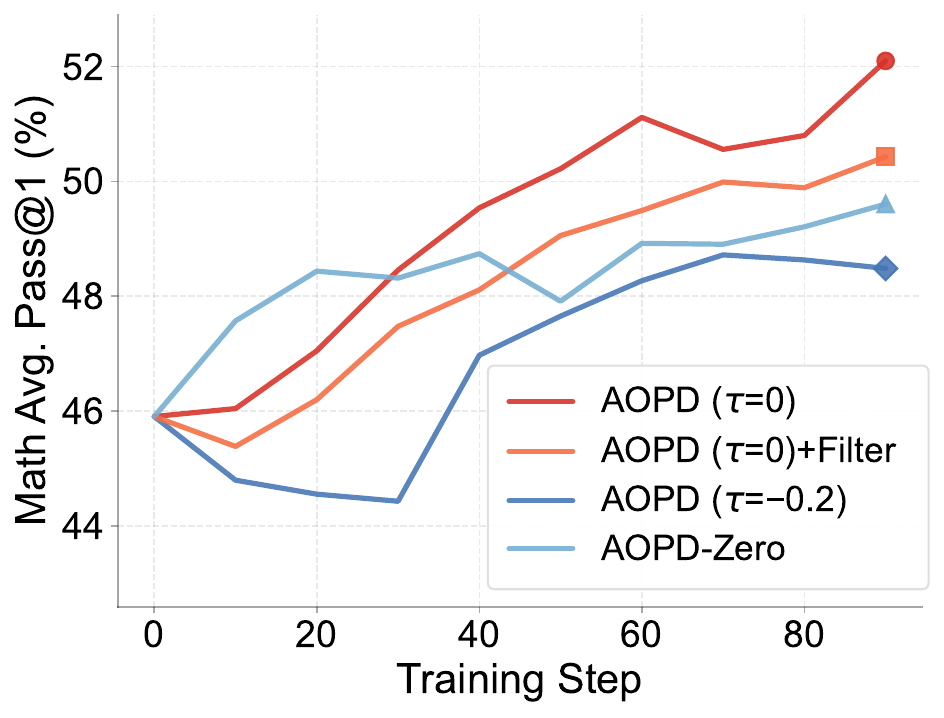}
        \caption{Effect of intervention locations.}
        \label{fig:ablation_tau_score}
\end{subfigure}
\caption{Ablation studies on top-$K$ and intervention location.}
\end{figure*}

\subsection{Ablation Study}
\subsubsection{Divergence Weighting in AOPD}
\label{sec:ablation_beta}
The parameter $\beta$ controls the interpolation between student-centric smoothing ($\beta \to 0$) and teacher-centric guidance ($\beta \to 1$) during interventions.
Figure~\ref{fig:ablation_beta} shows that AOPD attains its highest Pass@1 score at $\beta = 1$ and the highest Pass@4 at $\beta = 0.9$. 
More generally, the results show a clear trend that performance increases as the divergence becomes more biased toward the teacher distribution.
This empirical trend further validates the support-consistency principle discussed in Section~\ref{subsec:topk_and_divergence}.

\subsubsection{Effect of the Teacher Support Size}
\label{sec:ablation_k}
We further study the effect of the teacher support size $K$ in AOPD. 
As $K$ increases from 8 to 16 and 32 in Figure~\ref{fig:ablation_topk_bar}, the final Pass@1 scores on AIME 2024/2025 and HMMT 2025 (Feb) improve, suggesting that a larger top-$K$ preserves a richer portion of the teacher distribution and thus provides more complete signals. 
At the same time, we observe that smaller $K$ often yields larger gains at the early stage of training, as shown in Figure~\ref{fig:ablation_topk_aime25}.
This suggests that restricting supervision to a few high-probability teacher tokens produces denser and more concentrated learning signals, which is more beneficial for rapid initial improvement. 
In contrast, larger $K$ brings less pronounced early gains.
However, it retains a broader candidate distribution and therefore provides more informative signals in the later stage of training, leading to a higher final performance ceiling.
While larger $K$ could in principle further improve final performance, increasing $K$ beyond 32 only adds low-probability tokens with negligible training signal and degrades training efficiency. 
We therefore set $K=32$ as the default throughout our experiments.

\subsubsection{Effect of Intervention Location}
\label{sec:ablation_location}
The effectiveness of AOPD depends not only on whether KL guidance is used, but also on where the intervention is triggered. 
We consider three strategies: a token-level threshold $\tau$, which selects low-confidence positions for intervention; an accuracy filter, which applies KL guidance only to negative advantage regions from trajectories with incorrect final answers; and AOPD-Zero, which focuses exclusively on neutral regions with $A_t \approx 0$.

Figure~\ref{fig:ablation_tau_score} shows that standard AOPD with $\tau = 0$ achieves the strongest overall performance.
However, the more conservative setting $\tau = -0.2$ improves more slowly, indicating that shrinking the intervention range weakens the benefit of teacher guidance. 
Extended studies on $\tau$ are presented in Appendix~\ref{sec:ablation_tau}.
The accuracy-filtered setting remains competitive but consistently trails standard AOPD under the same threshold, suggesting that informative correction signals also arise in correct trajectories. 
Meanwhile, AOPD-Zero outperforms $\tau = -0.2$, which confirms that neutral regions are a major source of optimization stagnation in OPD.
These results highlight that the benefit of AOPD is not confined to correcting obvious errors, but also arises in correct trajectories or neutral positions.
Overall, the best results are achieved only by standard AOPD, and any reduction in the set of intervened positions leads to a drop in performance.

\section{Conclusion}
\label{sec:conclusion}
In this work, we identify three limitations of OPD and propose asymmetric on-policy distillation (AOPD).
AOPD preserves policy gradient updates where they are reliable and applies direct teacher distribution alignment where stronger correction is needed. 
Evaluations across multiple benchmarks show that AOPD consistently improves mathematical reasoning and remains stable under different model scales and initialization steps, where standard OPD often struggles. 
More broadly, our results suggest that combining exploitation and imitation at the token level can preserve policy entropy and better retain previously acquired capabilities.

\section*{Limitations}
\label{sec:limitations}
Our evaluation is limited to mathematical reasoning and tool-use settings, and it remains to be explored how AOPD performs on other task types.
Combining AOPD with reinforcement learning driven by verifiable rewards may yield additional gains on tasks where outcome-based reward is available.
In addition, the current instantiation of AOPD adopts a fixed intervention design, and future work may investigate more adaptive intervention strategies.

\newpage
\bibliography{custom}

@misc{qwen3,
      title={Qwen3 Technical Report}, 
      author={An Yang and Anfeng Li and Baosong Yang and Beichen Zhang and Binyuan Hui and Bo Zheng and Bowen Yu and Chang Gao and Chengen Huang and Chenxu Lv and Chujie Zheng and Dayiheng Liu and Fan Zhou and Fei Huang and Feng Hu and Hao Ge and Haoran Wei and Huan Lin and Jialong Tang and Jian Yang and Jianhong Tu and Jianwei Zhang and Jianxin Yang and Jiaxi Yang and Jing Zhou and Jingren Zhou and Junyang Lin and Kai Dang and Keqin Bao and Kexin Yang and Le Yu and Lianghao Deng and Mei Li and Mingfeng Xue and Mingze Li and Pei Zhang and Peng Wang and Qin Zhu and Rui Men and Ruize Gao and Shixuan Liu and Shuang Luo and Tianhao Li and Tianyi Tang and Wenbiao Yin and Xingzhang Ren and Xinyu Wang and Xinyu Zhang and Xuancheng Ren and Yang Fan and Yang Su and Yichang Zhang and Yinger Zhang and Yu Wan and Yuqiong Liu and Zekun Wang and Zeyu Cui and Zhenru Zhang and Zhipeng Zhou and Zihan Qiu},
      year={2025},
      eprint={2505.09388},
      archivePrefix={arXiv},
      primaryClass={cs.CL},
      url={https://arxiv.org/abs/2505.09388}, 
}

@misc{kimi25,
      title={Kimi K2.5: Visual Agentic Intelligence}, 
      author={{Kimi Team} and Tongtong Bai and Yifan Bai and Yiping Bao and S. H. Cai and Yuan Cao and Ziwei Chai and Y. Charles and H. S. Che and Cheng Chen and Guanduo Chen and Huarong Chen and Jia Chen and Jianlong Chen and Jun Chen and Kefan Chen and Liang Chen and Ruijue Chen and Xinhao Chen and Yanru Chen and Yanxu Chen and Yicun Chen and Yimin Chen and Yingjiang Chen and Yuankun Chen and Yujie Chen and Yutian Chen and Zhirong Chen and Ziwei Chen and Dazhi Cheng and Yean Cheng and Minghan Chu and Jialei Cui and Jiaqi Deng and Muxi Diao and Hao Ding and Mengfan Dong and Mengnan Dong and Yuxin Dong and Yuhao Dong and Angang Du and Chenzhuang Du and Dikang Du and Lingxiao Du and Yulun Du and Yu Fan and Shengjun Fang and Qiulin Feng and Yichen Feng and Garimugai Fu and Kelin Fu and Hongcheng Gao and Tong Gao and Yuyao Ge and Shangyi Geng and Chengyang Gong and Xiaochen Gong and Zhuoma Gongque and Qizheng Gu and Xinran Gu and Yicheng Gu and Longyu Guan and Shuhao Guan and Yuanying Guo and Xiaoru Hao and Dailan He and Tianhong He and Weiran He and Wenyang He and Yibo He and Yunjia He and Chao Hong and Hao Hu and Jiaxi Hu and Yangyang Hu and Zhenxing Hu and Ke Huang and Ruiyuan Huang and Weixiao Huang and Zhiqi Huang and Chaobo Jia and Tao Jiang and Zhejun Jiang and Xinyi Jin and Yu Jing and Guokun Lai and Aidi Li and C. Li and Cheng Li and Fang Li and Guanghe Li and Guanyu Li and Haitao Li and Haoyang Li and Jia Li and Jingwei Li and Junxiong Li and Lincan Li and Mo Li and Weihong Li and Wentao Li and Xinhang Li and Xinhao Li and Yang Li and Yanhao Li and Yiwei Li and Yuxiao Li and Zhaowei Li and Zhaoxi Li and Zheming Li and Weilong Liao and Jiawei Lin and Xiaohan Lin and Yibo Lin and Zhishan Lin and Zichao Lin and Cheng Liu and Chenyu Liu and Hongzhang Liu and Liang Liu and Shaowei Liu and Shudong Liu and Shuran Liu and Tianwei Liu and Tianyu Liu and Weizhou Liu and Xiangyan Liu and Yangyang Liu and Yanming Liu and Yibo Liu and Yuanxin Liu and Zhengying Liu and Zhongnuo Liu and Enzhe Lu and Haoyu Lu and Zhiyuan Lu and G. Luo and Junyu Luo and Tongxu Luo and Yashuo Luo and Long Ma and Shaoguang Mao and Yuan Mei and Xin Men and Fanqing Meng and Zhiyong Meng and Yibo Miao and Minqing Ni and Kun Ouyang and Siyuan Pan and Bo Pang and Yuchao Qian and Ruoyu Qin and Zeyu Qin and Jiezhong Qiu and Bowen Qu and Zeyu Shang and Youbo Shao and Tianxiao Shen and Zhennan Shen and Juanfeng Shi and Lidong Shi and Shengyuan Shi and Feifan Song and Pengwei Song and Tianhui Song and Xiaoxi Song and Hongjin Su and Jianlin Su and Zhaochen Su and Lin Sui and Jinsong Sun and Junyao Sun and Tongyu Sun and Flood Sung and Yunpeng Tai and Chuning Tang and Heyi Tang and Xiaojuan Tang and Zhengyang Tang and Jiawen Tao and Shiyuan Teng and Chaoran Tian and Pengfei Tian and Bowen Wang and Chensi Wang and Chuang Wang and Congcong Wang and Dingkun Wang and Dinglu Wang and Dongliang Wang and Feng Wang and Hailong Wang and Haiming Wang and Hao Wang and Hengzhi Wang and Huaqing Wang and Hui Wang and Jiahao Wang and Jinhong Wang and Jiuzheng Wang and Kaixin Wang and Linian Wang and Qibin Wang and Shengjie Wang and Shuyi Wang and Si Wang and Wei Wang and Xiaochen Wang and Xinyuan Wang and Yao Wang and Yejie Wang and Yipu Wang and Yiqin Wang and Yucheng Wang and Yuzhi Wang and Zhaoji Wang and Zhaowei Wang and Zhengtao Wang and Zhexu Wang and Zifan Wang and Zihan Wang and Zizhe Wang and Chu Wei and Ming Wei and Chuan Wen and Zichen Wen and Chengjie Wu and Haoning Wu and Junyan Wu and Rucong Wu and Wenhao Wu and Yuefeng Wu and Yuhao Wu and Yuxin Wu and Zijian Wu and Chenjun Xiao and Jin Xie and Xiaotong Xie and Yuchong Xie and Bowei Xing and Boyu Xu and Jianfan Xu and Jing Xu and Jinjing Xu and L. H. Xu and Lin Xu and Suting Xu and Weixin Xu and Xinbo Xu and Xinran Xu and Yangchuan Xu and Yichang Xu and Yuemeng Xu and Zelai Xu and Ziyao Xu and Junjie Yan and Yuzi Yan and Guangyao Yang and Hao Yang and Junwei Yang and Kai Yang and Ningyuan Yang and Xiaofei Yang and Xinlong Yang and Xinyu Yang and Ying Yang and Yi Yang and Yi Yang and Zhen Yang and Zhilin Yang and Zonghan Yang and Haotian Yao and Dan Ye and Haoran Ye and Wenjie Ye and Zhuorui Ye and Peng Yebo and Bohong Yin and Chengzhen Yu and Longhui Yu and Tao Yu and Tianxiang Yu and Enming Yuan and Mengjie Yuan and Xiaokun Yuan and Yang Yue and Weihao Zeng and Dunyuan Zha and Haobing Zhan and Dehao Zhang and Hao Zhang and Jin Zhang and Puqi Zhang and Qiao Zhang and Rui Zhang and Xiaobin Zhang and Xiaoyun Zhang and Y. Zhang and Yadong Zhang and Yangkun Zhang and Yichi Zhang and Yizhi Zhang and Yongting Zhang and Yu Zhang and Yushun Zhang and Yutao Zhang and Yutong Zhang and Zheng Zhang and Chenguang Zhao and Feifan Zhao and Jinxiang Zhao and Shuai Zhao and Xiangyu Zhao and Xuanle Zhao and Yikai Zhao and Zijia Zhao and Huabin Zheng and Ruihan Zheng and Shaojie Zheng and Tengyang Zheng and Junfeng Zhong and Longguang Zhong and Weiming Zhong and M. Zhou and Runjie Zhou and Xinyu Zhou and Zaida Zhou and Jinguo Zhu and Liya Zhu and Xinhao Zhu and Yuxuan Zhu and Zhen Zhu and Jingze Zhuang and Weiyu Zhuang and Ying Zou and Xinxing Zu},
      year={2026},
      eprint={2602.02276},
      archivePrefix={arXiv},
      primaryClass={cs.CL},
      url={https://arxiv.org/abs/2602.02276}, 
}

@inproceedings{
zhu2025surprising,
title={The Surprising Effectiveness of Negative Reinforcement in {LLM} Reasoning},
author={Xinyu Zhu and Mengzhou Xia and Zhepei Wei and Wei-Lin Chen and Danqi Chen and Yu Meng},
booktitle={The Thirty-ninth Annual Conference on Neural Information Processing Systems},
year={2025},
url={https://openreview.net/forum?id=ftVlLG9cks}
}

@inproceedings{tang2025samplepolarity,
title = {Rethinking Sample Polarity in Reinforcement Learning with Verifiable Rewards},
author = {Tang, Xinyu and
Zhan, Yuliang and
Li, Zhixun and
Zhao, Xin and
Zhang, Zhenduo and
Wen, Zujie and
Zhang, Zhiqiang and
Zhou, Jun},
booktitle = {Proceedings of the 64th Annual Meeting of the Association for Computational Linguistics (Volume 1: Long Papers)},
month = jul,
year = {2026},
address = {San Diego, California, United States},
publisher = {Association for Computational Linguistics},
url = {https://aclanthology.org/2026.acl-long.134/},
doi = {10.18653/v1/2026.acl-long.134},
pages = {2928--2954}
}

@misc{xu2024surveykd,
      title={A Survey on Knowledge Distillation of Large Language Models}, 
      author={Xiaohan Xu and Ming Li and Chongyang Tao and Tao Shen and Reynold Cheng and Jinyang Li and Can Xu and Dacheng Tao and Tianyi Zhou},
      year={2024},
      eprint={2402.13116},
      archivePrefix={arXiv},
      primaryClass={cs.CL},
      url={https://arxiv.org/abs/2402.13116}, 
}

@misc{hinton2015distilling,
      title={Distilling the Knowledge in a Neural Network}, 
      author={Geoffrey Hinton and Oriol Vinyals and Jeff Dean},
      year={2015},
      eprint={1503.02531},
      archivePrefix={arXiv},
      primaryClass={stat.ML},
      url={https://arxiv.org/abs/1503.02531}, 
}

@inproceedings{distilling2022stepbystep,
title = {Distilling Step-by-Step! Outperforming Larger Language Models with Less Training Data and Smaller Model Sizes},
author = {Hsieh, Cheng-Yu and
Li, Chun-Liang and
Yeh, Chih-kuan and
Nakhost, Hootan and
Fujii, Yasuhisa and
Ratner, Alex and
Krishna, Ranjay and
Lee, Chen-Yu and
Pfister, Tomas},
booktitle = {Findings of the Association for Computational Linguistics: ACL 2023},
month = jul,
year = {2023},
address = {Toronto, Canada},
publisher = {Association for Computational Linguistics},
url = {https://aclanthology.org/2023.findings-acl.507/},
doi = {10.18653/v1/2023.findings-acl.507},
pages = {8003--8017}
}

@inproceedings{han2024tinyllm,
author = {Tian, Yijun and
Han, Yikun and
Chen, Xiusi and
Wang, Wei and
Chawla, Nitesh V.},
title = {Beyond Answers: Transferring Reasoning Capabilities to Smaller {LLM}s Using Multi-Teacher Knowledge Distillation},
booktitle = {Proceedings of the Eighteenth ACM International Conference on Web Search and Data Mining},
year = {2025},
pages = {251--260},
publisher = {Association for Computing Machinery},
address = {New York, NY, USA},
location = {Hannover, Germany},
series = {WSDM '25},
doi = {10.1145/3701551.3703577},
url = {https://doi.org/10.1145/3701551.3703577}
}

@inproceedings{
gu2023minillm,
title={Mini{LLM}: Knowledge Distillation of Large Language Models},
author={Yuxian Gu and Li Dong and Furu Wei and Minlie Huang},
booktitle={The Twelfth International Conference on Learning Representations},
year={2024},
url={https://openreview.net/forum?id=5h0qf7IBZZ}
}

@inproceedings{
agarwal2024gkd,
title={On-Policy Distillation of Language Models: Learning from Self-Generated Mistakes},
author={Rishabh Agarwal and Nino Vieillard and Yongchao Zhou and Piotr Stanczyk and Sabela Ramos Garea and Matthieu Geist and Olivier Bachem},
booktitle={The Twelfth International Conference on Learning Representations},
year={2024},
url={https://openreview.net/forum?id=3zKtaqxLhW}
}

@article{thinkingmachines_opd, 
  author    = {Lu, Kevin and {Thinking Machines Lab}}, 
  title     = {{O}n-{P}olicy Distillation}, 
  journal   = {Thinking Machines Lab: Connectionism}, 
  year      = {2025}, 
  doi       = {10.64434/tml.20251026},
  url       = {https://thinkingmachines.ai/blog/on-policy-distillation}
}

@inproceedings{
shenfeld2025sdft,
title={Self-Distillation Enables Continual Learning},
author={Idan Shenfeld and Mehul Damani and Jonas H{\"u}botter and Pulkit Agrawal},
booktitle={Forty-third International Conference on Machine Learning},
year={2026},
url={https://openreview.net/forum?id=qA6FgH0nnZ}
}

@inproceedings{
zhao2026opsd,
title={Self-Distilled Reasoner: On-Policy Self-Distillation for Large Language Models},
author={Siyan Zhao and Zhihui Xie and Mengchen Liu and Jing Huang and Guan Pang and Feiyu Chen and Aditya Grover},
booktitle={Forty-third International Conference on Machine Learning},
year={2026},
url={https://openreview.net/forum?id=Jpxfof0EaS}
}

@misc{zhang2026opsdl,
      title={OPSDL: On-Policy Self-Distillation for Long-Context Language Models}, 
      author={Xinsen Zhang and Zhenkai Ding and Tianjun Pan and Run Yang and Chun Kang and Xue Xiong and Jingnan Gu},
      year={2026},
      eprint={2604.17535},
      archivePrefix={arXiv},
      primaryClass={cs.CL},
      url={https://arxiv.org/abs/2604.17535}, 
}

@misc{zheng2026scope,
      title={SCOPE: Signal-Calibrated On-Policy Distillation Enhancement with Dual-Path Adaptive Weighting}, 
      author={Binbin Zheng and Xing Ma and Yiheng Liang and Jingqing Ruan and Xiaoliang Fu and Kepeng Lin and Benchang Zhu and Ke Zeng and Xunliang Cai},
      year={2026},
      eprint={2604.10688},
      archivePrefix={arXiv},
      primaryClass={cs.LG},
      url={https://arxiv.org/abs/2604.10688}, 
}

@misc{li2026rethinkopd,
      title={Rethinking On-Policy Distillation of Large Language Models: Phenomenology, Mechanism, and Recipe}, 
      author={Yaxuan Li and Yuxin Zuo and Bingxiang He and Jinqian Zhang and Chaojun Xiao and Cheng Qian and Tianyu Yu and Huan-ang Gao and Wenkai Yang and Zhiyuan Liu and Ning Ding},
      year={2026},
      eprint={2604.13016},
      archivePrefix={arXiv},
      primaryClass={cs.LG},
      url={https://arxiv.org/abs/2604.13016}, 
}

@misc{gu2026copd,
      title={Co-Evolving Policy Distillation}, 
      author={Naibin Gu and Chenxu Yang and Qingyi Si and Chuanyu Qin and Dingyu Yao and Peng Fu and Zheng Lin and Weiping Wang and Nan Duan and Jiaqi Wang},
      year={2026},
      eprint={2604.27083},
      archivePrefix={arXiv},
      primaryClass={cs.LG},
      url={https://arxiv.org/abs/2604.27083}, 
}

@misc{mimov2,
      title={MiMo-V2-Flash Technical Report}, 
      author={{Xiaomi LLM-Core Team} and Bangjun Xiao and Bingquan Xia and Bo Yang and Bofei Gao and Bowen Shen and Chen Zhang and Chenhong He and Chiheng Lou and Fuli Luo and Gang Wang and Gang Xie and Hailin Zhang and Hanglong Lv and Hanyu Li and Heyu Chen and Hongshen Xu and Houbin Zhang and Huaqiu Liu and Jiangshan Duo and Jianyu Wei and Jiebao Xiao and Jinhao Dong and Jun Shi and Junhao Hu and Kainan Bao and Kang Zhou and Lei Li and Liang Zhao and Linghao Zhang and Peidian Li and Qianli Chen and Shaohui Liu and Shihua Yu and Shijie Cao and Shimao Chen and Shouqiu Yu and Shuo Liu and Tianling Zhou and Weijiang Su and Weikun Wang and Wenhan Ma and Xiangwei Deng and Bohan Mao and Bowen Ye and Can Cai and Chenghua Wang and Chengxuan Zhu and Chong Ma and Chun Chen and Chunan Li and Dawei Zhu and Deshan Xiao and Dong Zhang and Duo Zhang and Fangyue Liu and Feiyu Yang and Fengyuan Shi and Guoan Wang and Hao Tian and Hao Wu and Heng Qu and Hongfei Yi and Hongxu An and Hongyi Guan and Xing Zhang and Yifan Song and Yihan Yan and Yihao Zhao and Yingchun Lai and Yizhao Gao and Yu Cheng and Yuanyuan Tian and Yudong Wang and Zhen Tang and Zhengju Tang and Zhengtao Wen and Zhichao Song and Zhixian Zheng and Zihan Jiang and Jian Wen and Jiarui Sun and Jiawei Li and Jinlong Xue and Jun Xia and Kai Fang and Menghang Zhu and Nuo Chen and Qian Tu and Qihao Zhang and Qiying Wang and Rang Li and Rui Ma and Shaolei Zhang and Shengfan Wang and Shicheng Li and Shuhao Gu and Shuhuai Ren and Sirui Deng and Tao Guo and Tianyang Lu and Weiji Zhuang and Weikang Zhang and Weimin Xiong and Wenshan Huang and Wenyu Yang and Xin Zhang and Xing Yong and Xu Wang and Xueyang Xie and Yilin Jiang and Yixin Yang and Yongzhe He and Yu Tu and Yuanliang Dong and Yuchen Liu and Yue Ma and Yue Yu and Yuxing Xiang and Zhaojun Huang and Zhenru Lin and Zhipeng Xu and Zhiyang Chen and Zhonghua Deng and Zihan Zhang and Zihao Yue},
      year={2026},
      eprint={2601.02780},
      archivePrefix={arXiv},
      primaryClass={cs.CL},
      url={https://arxiv.org/abs/2601.02780}, 
}

@misc{deepseekv4,
      title={DeepSeek-V4: Towards Highly Efficient Million-Token Context Intelligence}, 
      author={DeepSeek-AI and Anyi Xu and Bangcai Lin and Bing Xue and Bingxuan Wang and Bingzheng Xu and Bochao Wu and Bowei Zhang and Chaofan Lin and Chen Dong and Chenchen Ling and Chengda Lu and Chenggang Zhao and Chengqi Deng and Chengyu Hou and Chenhao Xu and Chenze Shao and Chong Ruan and Conner Sun and Damai Dai and Daya Guo and Dejian Yang and Deli Chen and Donghao Li and Dongjie Ji and Erhang Li and Fang Wei and Fangyun Lin and Fangzhou Yuan and Feiyu Xia and Fucong Dai and Guangbo Hao and Guanting Chen and Guoai Cao and Guolai Meng and Guowei Li and Han Yu and Han Zhang and Hanwei Xu and Hao Li and Haofen Liang and Haoling Zhang and Haoming Luo and Haoran Wei and Haotian Yuan and Haowei Zhang and Haowen Luo and Haoyu Chen and Haozhe Ji and Hengqing Zhang and Honghui Ding and Hongxuan Tang and Huanqi Cao and Huazuo Gao and Hui Qu and Hui Zeng and J Yang and JQ Zhu and Jia Luo and Jia Song and Jia Yu and Jialiang Huang and Jialu Cai and Jian Liang and Jiangting Zhou and Jiasheng Ye and Jiashi Li and Jiaxin Xu and Jiewen Hu and Jieyu Yang and Jin Chen and Jin Yan and Jingchang Chen and Jingli Zhou and Jingting Xiang and Jingyang Yuan and Jingyuan Cheng and Jingzi Zhou and Jinhua Zhu and Jiping Yu and Joseph Sun and Jun Ran and Junguang Jiang and Junjie Qiu and Junlong Li and Junmin Zheng and Junxiao Song and Kai Dong and Kaige Gao and Kang Guan and Kexing Zhou and Kezhao Huang and Kuai Yu and Lean Wang and Lecong Zhang and Lei Wang and Leyi Xia and Li Zhang and Liang Zhao and Lihua Guo and Lingxiao Luo and Linwang Ma and Linyan Zhu and Litong Wang and Liyu Cai and Liyue Zhang and Longhao Chen and MS Di and MY Xu and Max Mei and Miaojun Wang and Mingchuan Zhang and Minghua Zhang and Minghui Tang and Mingming Li and Mingxu Zhou and Minmin Han and Ning Wang and Panpan Huang and Panpan Wang and Peixin Cong and Peiyi Wang and Peng Zhang and Qiancheng Wang and Qihao Zhu and Qingyang Li and Qinyu Chen and Qiushi Du and Qiwei Jiang and Rui Tian and Ruifan Xu and Ruijie Lu and Ruiling Xu and Ruiqi Ge and Ruisong Zhang and Ruizhe Pan and Runji Wang and Runqian Chen and Runqiu Yin and Runxin Xu and Ruomeng Shen and Ruoyu Zhang and Ruyi Chen and SH Liu and Shanghao Lu and Shangmian Sun and Shangyan Zhou and Shanhuang Chen and Shaofei Cai and Shaoheng Nie and Shaoqing Wu and Shaoyuan Chen and Shengding Hu and Shengyu Liu and Shiqiang Hu and Shirong Ma and Shiyu Wang and Shuiping Yu and Shunfeng Zhou and Shuting Pan and Shuying Yu and Songyang Zhou and Tao Ni and Tao Yun and Tian Jin and Tian Pei and Tian Ye and Tianle Lin and Tianran Ji and Tianyi Cui and Tianyuan Yue and Tingting Yu and Tun Wang and W Zhang and WL Xiao and Wangding Zeng and Wei An and Weilin Zhao and Wen Liu and Wenfeng Liang and Wenjie Pang and Wenjing Luo and Wenjing Yao and Wenjun Gao and Wenkai Yang and Wenlve Huang and Wenqing Hou and Wentao Zhang and Wenting Ma and Xi Gao and Xiang He and Xiangwen Wang and Xianzu Wang and Xiao Bi and Xiaodong Liu and Xiaohan Wang and Xiaokang Chen and Xiaokang Zhang and Xiaotao Nie and Xiaowen Sun and Xiaoxiang Wang and Xin Cheng and Xin Liu and Xin Xie and Xingchao Liu and Xingchen Liu and Xingkai Yu and Xingyou Li and Xinyu Yang and Xinyu Zhang and Xu Chen and Xuanyu Wang and Xuecheng Su and Xueyin Chen and Xuheng Lin and Xuwei Fu and YC Yan and YQ Wang and YW Ma and Yanfeng Luo and Yang Zhang and Yanhong Xu and Yanru Ma and Yanwen Huang and Yao Li and Yao Li and Yao Xu and Yao Zhao and Yaofeng Sun and Yaohui Wang and Yi Qian and Yi Shao and Yi Yu and Yichao Zhang and Yifan Ding and Yifan Shi and Yijia Wu and Yiliang Xiong and Yiling Ma and Ying He and Ying Tang and Ying Zhou and Yingjia Luo and Yinmin Zhong and Yishi Piao and Yisong Wang and Yixiang Zhang and Yixiao Chen and Yixuan Tan and Yixuan Wei and Yiyang Ma and Yiyuan Liu and Yonglun Yang and Yongqiang Guo and Yongtong Wu and Yu Wu and YuKun Li and Yuan Cheng and Yuan Ou and Yuanfan Xu and Yuanhao Li and Yuduan Wang and Yuehan Yang and Yuer Xu and Yuhan Wu and Yuhao Meng and Yuheng Zou and Yukun Zha and Yunfan Xiong and Yupeng Chen and Yuping Lin and Yuqian Cao and Yuqian Wang and Yushun Zhang and Yuting Yan and Yutong Lin and Yuxian Gu and Yuxiang Luo and Yuxiang You and Yuxuan Liu and Yuxuan Zhou and Yuyang Zhou and Yuzhen Huang and ZF Wu and Zehao Wang and Zehua Zhao and Zehui Ren and Zekai Zhang and Zhangli Sha and Zhe Fu and Zhe Ju and Zhean Xu and Zhenda Xie and Zhengyan Zhang and Zheren Gao and Zhewen Hao and Zhibin Gou and Zhicheng Ma and Zhigang Yan and Zhihong Shao and Zhixian Huang and Zhixuan Chen and Zhiyu Wu and Zhizhou Ren and Zhongyu Wu and Zhuoshu Li and Zhuping Zhang and Zian Xu and Zihao Wang and Zihua Qu and Zihui Gu and Zijia Zhu and Zilin Li and Zipeng Zhang and Ziwei Xie and Ziyi Gao and Ziyi Wan and Zizheng Pan and Zongqing Yao},
      year={2026},
      eprint={2606.19348},
      archivePrefix={arXiv},
      primaryClass={cs.CL},
      url={https://arxiv.org/abs/2606.19348}, 
}

@misc{glm5,
      title={GLM-5: from Vibe Coding to Agentic Engineering}, 
      author={{GLM-5-Team} and Aohan Zeng and Xin Lv and Zhenyu Hou and Zhengxiao Du and Qinkai Zheng and Bin Chen and Da Yin and Chendi Ge and Chenghua Huang and Chengxing Xie and Chenzheng Zhu and Congfeng Yin and Cunxiang Wang and Gengzheng Pan and Hao Zeng and Haoke Zhang and Haoran Wang and Huilong Chen and Jiajie Zhang and Jian Jiao and Jiaqi Guo and Jingsen Wang and Jingzhao Du and Jinzhu Wu and Kedong Wang and Lei Li and Lin Fan and Lucen Zhong and Mingdao Liu and Mingming Zhao and Pengfan Du and Qian Dong and Rui Lu and Shuang-Li and Shulin Cao and Song Liu and Ting Jiang and Xiaodong Chen and Xiaohan Zhang and Xuancheng Huang and Xuezhen Dong and Yabo Xu and Yao Wei and Yifan An and Yilin Niu and Yitong Zhu and Yuanhao Wen and Yukuo Cen and Yushi Bai and Zhongpei Qiao and Zihan Wang and Zikang Wang and Zilin Zhu and Ziqiang Liu and Zixuan Li and Bojie Wang and Bosi Wen and Can Huang and Changpeng Cai and Chao Yu and Chen Li and Chengwei Hu and Chenhui Zhang and Dan Zhang and Daoyan Lin and Dayong Yang and Di Wang and Ding Ai and Erle Zhu and Fangzhou Yi and Feiyu Chen and Guohong Wen and Hailong Sun and Haisha Zhao and Haiyi Hu and Hanchen Zhang and Hanrui Liu and Hanyu Zhang and Hao Peng and Hao Tai and Haobo Zhang and He Liu and Hongwei Wang and Hongxi Yan and Hongyu Ge and Huan Liu and Huanpeng Chu and Jia'ni Zhao and Jiachen Wang and Jiajing Zhao and Jiamin Ren and Jiapeng Wang and Jiaxin Zhang and Jiayi Gui and Jiayue Zhao and Jijie Li and Jing An and Jing Li and Jingwei Yuan and Jinhua Du and Jinxin Liu and Junkai Zhi and Junwen Duan and Kaiyue Zhou and Kangjian Wei and Ke Wang and Keyun Luo and Laiqiang Zhang and Leigang Sha and Liang Xu and Lindong Wu and Lintao Ding and Lu Chen and Minghao Li and Nianyi Lin and Pan Ta and Qiang Zou and Rongjun Song and Ruiqi Yang and Shangqing Tu and Shangtong Yang and Shaoxiang Wu and Shengyan Zhang and Shijie Li and Shuang Li and Shuyi Fan and Wei Qin and Wei Tian and Weining Zhang and Wenbo Yu and Wenjie Liang and Xiang Kuang and Xiangmeng Cheng and Xiangyang Li and Xiaoquan Yan and Xiaowei Hu and Xiaoying Ling and Xing Fan and Xingye Xia and Xinyuan Zhang and Xinze Zhang and Xirui Pan and Xu Zou and Xunkai Zhang and Yadi Liu and Yandong Wu and Yanfu Li and Yidong Wang and Yifan Zhu and Yijun Tan and Yilin Zhou and Yiming Pan and Ying Zhang and Yinpei Su and Yipeng Geng and Yong Yan and Yonglin Tan and Yuean Bi and Yuhan Shen and Yuhao Yang and Yujiang Li and Yunan Liu and Yunqing Wang and Yuntao Li and Yurong Wu and Yutao Zhang and Yuxi Duan and Yuxuan Zhang and Zezhen Liu and Zhengtao Jiang and Zhenhe Yan and Zheyu Zhang and Zhixiang Wei and Zhuo Chen and Zhuoer Feng and Zijun Yao and Ziwei Chai and Ziyuan Wang and Zuzhou Zhang and Bin Xu and Minlie Huang and Hongning Wang and Juanzi Li and Yuxiao Dong and Jie Tang},
      year={2026},
      eprint={2602.15763},
      archivePrefix={arXiv},
      primaryClass={cs.LG},
      url={https://arxiv.org/abs/2602.15763}, 
}

@inproceedings{bengio2015scheduled,
author = {Bengio, Samy and
Vinyals, Oriol and
Jaitly, Navdeep and
Shazeer, Noam},
title = {Scheduled Sampling for Sequence Prediction with Recurrent Neural Networks},
booktitle = {Advances in Neural Information Processing Systems},
volume = {28},
year = {2015},
pages = {1171--1179},
publisher = {MIT Press},
address = {Cambridge, MA, USA}
}

@inproceedings{
deepmath2025,
title={DeepMath-103K: A Large-Scale, Challenging, Decontaminated, and Verifiable Mathematical Dataset for Advancing Reasoning},
author={Zhiwei He and Tian Liang and Jiahao Xu and Qiuzhi Liu and Xingyu Chen and Yue Wang and Linfeng Song and Dian Yu and Zhenwen Liang and Wenxuan Wang and Zhuosheng Zhang and Rui Wang and Zhaopeng Tu and Haitao Mi and Dong Yu},
booktitle={The Fourteenth International Conference on Learning Representations},
year={2026},
url={https://openreview.net/forum?id=kHB5Te5IWm}
}

@inproceedings{kim2016sequence,
title = {Sequence-Level Knowledge Distillation},
author = {Kim, Yoon and
Rush, Alexander M.},
booktitle = {Proceedings of the 2016 Conference on Empirical Methods in Natural Language Processing},
month = nov,
year = {2016},
address = {Austin, Texas},
publisher = {Association for Computational Linguistics},
url = {https://aclanthology.org/D16-1139/},
doi = {10.18653/v1/D16-1139},
pages = {1317--1327}
}

@misc{yang2026learning,
      title={Learning beyond Teacher: Generalized On-Policy Distillation with Reward Extrapolation}, 
      author={Wenkai Yang and Weijie Liu and Ruobing Xie and Kai Yang and Saiyong Yang and Yankai Lin},
      year={2026},
      eprint={2602.12125},
      archivePrefix={arXiv},
      primaryClass={cs.LG},
      url={https://arxiv.org/abs/2602.12125}, 
}

@inproceedings{
yan2025luffy,
title={Learning to Reason under Off-Policy Guidance},
author={Jianhao Yan and Yafu Li and Zican Hu and Zhi Wang and Ganqu Cui and Xiaoye Qu and Yu Cheng and Yue Zhang},
booktitle={The Thirty-ninth Annual Conference on Neural Information Processing Systems},
year={2025},
url={https://openreview.net/forum?id=vO8LLoNWWk}
}

@inproceedings{ma2025relift,
title={Learning What Reinforcement Learning Can't: Interleaved Online Fine-Tuning for Hardest Questions},
author={Lu Ma and Hao Liang and Meiyi Qiang and Lexiang Tang and Xiaochen Ma and Zhen Hao Wong and Junbo Niu and Chengyu Shen and Runming He and Yanhao Li and Bin Cui and Wentao Zhang},
booktitle={The Fourteenth International Conference on Learning Representations},
year={2026},
url={https://openreview.net/forum?id=LzCBLrNoyM}
}

@inproceedings{
chord2025,
title={On-Policy {RL} Meets Off-Policy Experts: Harmonizing Supervised Fine-Tuning and Reinforcement Learning via Dynamic Weighting},
author={Wenhao Zhang and Yuexiang Xie and Yuchang Sun and Yanxi Chen and Guoyin Wang and Yaliang Li and Bolin Ding and Jingren Zhou},
booktitle={The Fourteenth International Conference on Learning Representations},
year={2026},
url={https://openreview.net/forum?id=dCm9bBrk5d}
}

@inproceedings{
chen2025beyond,
title={Beyond Two-Stage Training: Cooperative {SFT} and {RL} for {LLM} Reasoning},
author={Liang Chen and Xueting Han and Li Shen and Jing Bai and Kam-Fai Wong},
booktitle={Forty-third International Conference on Machine Learning},
year={2026},
url={https://openreview.net/forum?id=xWvj03N4sJ}
}

@inproceedings{
fu2025srft,
title={{SRFT}: A Single-Stage Method with Supervised and Reinforcement Fine-Tuning for Reasoning},
author={Yuqian Fu and Tinghong Chen and Jiajun Chai and Xihuai Wang and Songjun Tu and Guojun Yin and Wei Lin and Qichao Zhang and Yuanheng Zhu and Dongbin Zhao},
booktitle={The Fourteenth International Conference on Learning Representations},
year={2026},
url={https://openreview.net/forum?id=n6E0r6kQWQ}
}

@inproceedings{
huang2025blending,
title={Blending Supervised and Reinforcement Fine-Tuning with Prefix Sampling},
author={Zeyu Huang and Tianhao Cheng and Zihan Qiu and Zili Wang and Yinghui Xu and Edoardo M. Ponti and Ivan Titov},
booktitle={Forty-third International Conference on Machine Learning},
year={2026},
url={https://openreview.net/forum?id=MgtyPYrC4h}
}

@misc{chen2025stepwise,
      title={Step-wise Adaptive Integration of Supervised Fine-tuning and Reinforcement Learning for Task-Specific LLMs}, 
      author={Jack Chen and Fazhong Liu and Naruto Liu and Yuhan Luo and Erqu Qin and Harry Zheng and Tian Dong and Haojin Zhu and Yan Meng and Xiao Wang},
      year={2025},
      eprint={2505.13026},
      archivePrefix={arXiv},
      primaryClass={cs.LG},
      url={https://arxiv.org/abs/2505.13026}, 
}

@inproceedings{
guha2025openthoughts,
title={OpenThoughts: Data Recipes for Reasoning Models},
author={Etash Kumar Guha and Ryan Marten and Sedrick Keh and Negin Raoof and Georgios Smyrnis and Hritik Bansal and Marianna Nezhurina and Jean Mercat and Trung Vu and Zayne Rea Sprague and Ashima Suvarna and Benjamin Feuer and Leon Liangyu Chen and Zaid Khan and Eric Frankel and Sachin Grover and Caroline Choi and Niklas Muennighoff and Shiye Su and Wanjia Zhao and John Yang and Shreyas Pimpalgaonkar and Kartik sharma and Charlie Cheng-Jie Ji and Yichuan Deng and Sarah M Pratt and Vivek Ramanujan and Jon Saad-Falcon and Stutee Acharya and Jeffrey Li and Achal Dave and Alon Albalak and Kushal Arora and Blake Wulfe and Chinmay Hegde and Greg Durrett and Sewoong Oh and Mohit Bansal and Saadia Gabriel and Aditya Grover and Kai-Wei Chang and Vaishaal Shankar and Aaron Gokaslan and Mike A Merrill and Tatsunori Hashimoto and Yejin Choi and Jenia Jitsev and Reinhard Heckel and Maheswaran Sathiamoorthy and Alex Dimakis and Ludwig Schmidt},
booktitle={The Fourteenth International Conference on Learning Representations},
year={2026},
url={https://openreview.net/forum?id=7xjoTuaNmN}
}

@article{tang2023toolalpaca,
  author  = {Tang, Qiaoyu and Deng, Ziliang and Lin, Hongyu and Han, Xianpei and Liang, Qiao and Cao, Boxi and Sun, Le},
  title   = {{ToolAlpaca}: Generalized Tool Learning for Language Models with 3000 Simulated Cases},
  journal = {arXiv preprint arXiv:2306.05301},
  year    = {2023},
  url     = {https://arxiv.org/abs/2306.05301}
}

@inproceedings{sheng2024hybridflow, 
   title={HybridFlow: A Flexible and Efficient RLHF Framework},
   url={http://dx.doi.org/10.1145/3689031.3696075},
   DOI={10.1145/3689031.3696075},
   booktitle={Proceedings of the Twentieth European Conference on Computer Systems},
   publisher={ACM},
   author={Sheng, Guangming and Zhang, Chi and Ye, Zilingfeng and Wu, Xibin and Zhang, Wang and Zhang, Ru and Peng, Yanghua and Lin, Haibin and Wu, Chuan},
   year={2025},
   month=Mar, pages={1279–1297},
   collection={EuroSys ’25} }

@inproceedings{kwon2023efficient,
author = {Kwon, Woosuk and
Li, Zhuohan and
Zhuang, Siyuan and
Sheng, Ying and
Zheng, Lianmin and
Yu, Cody Hao and
Gonzalez, Joseph and
Zhang, Hao and
Stoica, Ion},
title = {Efficient Memory Management for Large Language Model Serving with PagedAttention},
booktitle = {Proceedings of the 29th Symposium on Operating Systems Principles},
year = {2023},
pages = {611--626},
publisher = {Association for Computing Machinery},
address = {New York, NY, USA},
location = {Koblenz, Germany},
series = {SOSP '23},
doi = {10.1145/3600006.3613165},
url = {https://doi.org/10.1145/3600006.3613165}
}

@inproceedings{
eopd,
title={Entropy-Aware On-Policy Distillation of Language Models},
author={Woogyeol Jin and Taywon Min and Yongjin Yang and Dennis Wei and Yi Zhou and Swanand Ravindra Kadhe and Nathalie Baracaldo and Kimin Lee},
booktitle={Forty-third International Conference on Machine Learning},
year={2026},
url={https://openreview.net/forum?id=J5i09faOOf}
}

@misc{kdrl,
      title={KDRL: Post-Training Reasoning LLMs via Unified Knowledge Distillation and Reinforcement Learning}, 
      author={Hongling Xu and Qi Zhu and Heyuan Deng and Jinpeng Li and Lu Hou and Yasheng Wang and Lifeng Shang and Ruifeng Xu and Fei Mi},
      year={2025},
      eprint={2506.02208},
      archivePrefix={arXiv},
      primaryClass={cs.LG},
      url={https://arxiv.org/abs/2506.02208}, 
}

\newpage
\appendix
\raggedbottom
\section{Related Work about Joint SFT-RL Optimization}
\label{sec:appendix_related_work}
Because on-policy distillation follows a reinforcement learning paradigm, it remains vulnerable to exploration bottlenecks. 
To alleviate this issue, recent work incorporates supervised learning or expert demonstrations into RL. 
Some methods dynamically interleave SFT and RL, such as ReLIFT~\citep{ma2025relift}, SASR~\citep{chen2025stepwise}, and BRIDGE~\citep{chen2025beyond}, while others unify them in a single-stage objective, including SRFT~\citep{fu2025srft} and Prefix-RFT~\citep{huang2025blending}. 
A related line instead interprets SFT as off-policy guidance within on-policy RL, as in LUFFY~\citep{yan2025luffy} and CHORD~\citep{chord2025}.
In this spirit, AOPD introduces localized soft-label supervised learning to improve exploration efficiency in OPD.

\section{Extended Analysis on Truncated Teacher Guidance}
\label{sec:appendix_fkl}

\begin{figure*}[h]
\centering
\begin{subfigure}[t]{0.32\textwidth}
    \centering
    \includegraphics[width=\linewidth]{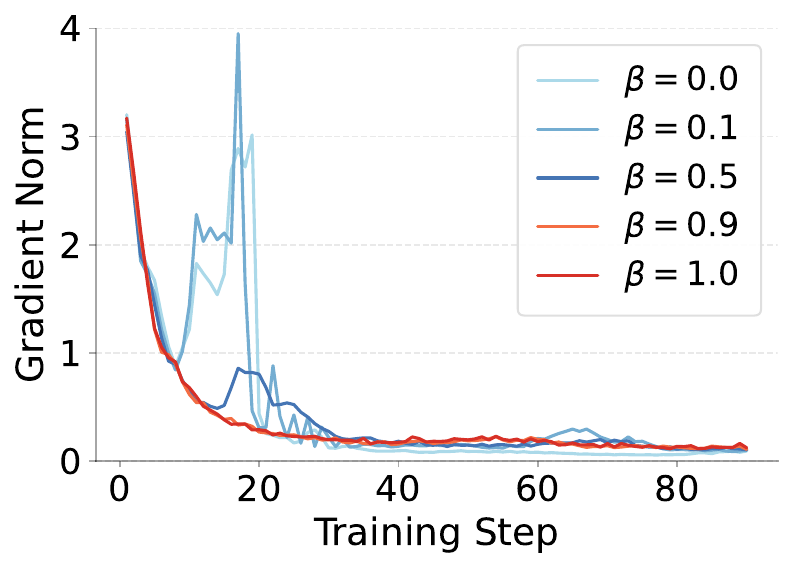}
    \caption{Gradient norm during training.}
    \label{fig:beta_grad_norm}
\end{subfigure}
\hfill
\begin{subfigure}[t]{0.32\textwidth}
    \centering
    \includegraphics[width=\linewidth]{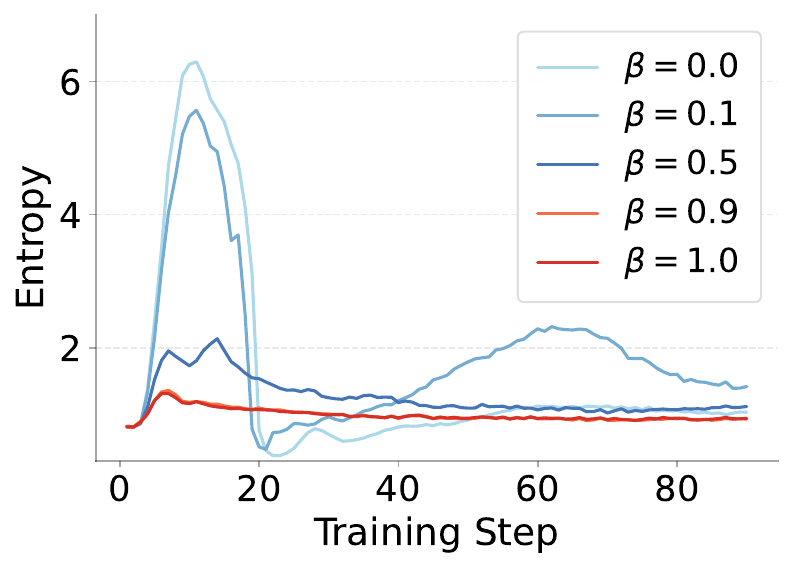}
    \caption{Policy entropy evolution.}
    \label{fig:beta_entropy}
\end{subfigure}
\hfill
\begin{subfigure}[t]{0.32\textwidth}
    \centering
    \includegraphics[width=\linewidth]{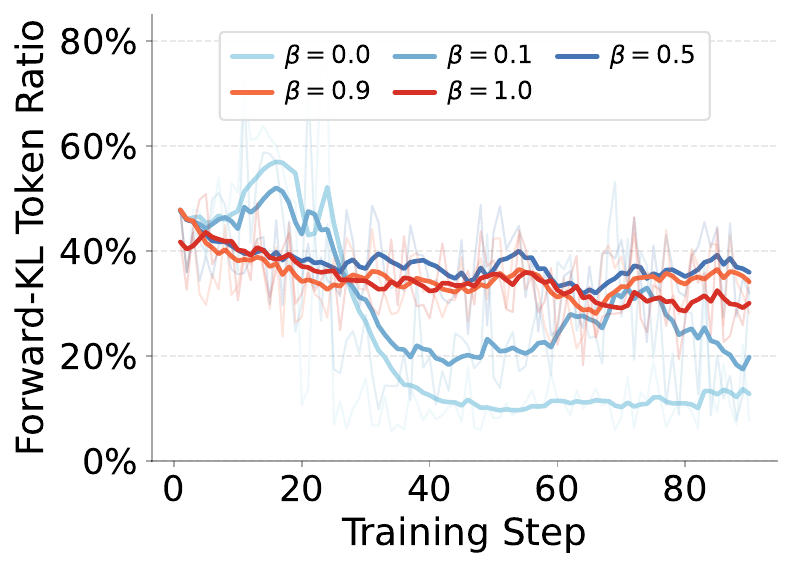}
    \caption{Ratios of tokens receiving JS divergence guidance.}
    \label{fig:beta_token_ratio}
\end{subfigure}
\caption{Training dynamics under different $\beta$ values.}
\label{fig:beta_dynamics}
\end{figure*}

\subsection{Gradient Analysis}
\label{sec:appendix_fkl_gradient}
In this section, we compare the logit-level gradients induced by forward KL and reverse KL under the same teacher-defined support $S_t$.
For simplicity, $P_S(\cdot \mid c_t)$ and $P_T(\cdot \mid c_t)$ denote the normalized student and teacher distributions on $S_t$.
We first consider the forward KL objective
\begin{equation}
\begin{split}
\mathcal{L}^{\mathrm{FKL}}_t
&= D_{\mathrm{KL}}\!\left(P_T(\cdot \mid c_t) \,\|\, P_S(\cdot \mid c_t)\right) \\
&= \sum_{v \in S_t} P_T(v \mid c_t) \log \frac{P_T(v \mid c_t)}{P_S(v \mid c_t)}.
\end{split}
\end{equation}
Taking derivatives with respect to the student logits, the term
$\sum_{v \in S_t} P_T(v \mid c_t)\log P_T(v \mid c_t)$
can be treated as a constant.
We therefore have
\begin{equation}
\begin{split}
\frac{\partial \mathcal{L}^{\mathrm{FKL}}_t}{\partial z_j}
&= \frac{\partial}{\partial z_j}\left(-\sum_{v \in S_t}P_T(v \mid c_t)\log P_S(v \mid c_t)\right) \\
&= P_S(j \mid c_t)-P_T(j \mid c_t),
\qquad j \in S_t.
\end{split}
\label{eq:appendixforwardgrad}
\end{equation}
We next consider the reverse KL objective
\begin{equation}
\begin{split}
\mathcal{L}^{\mathrm{RKL}}_t
&= D_{\mathrm{KL}}\!\left(P_S(\cdot \mid c_t) \,\|\, P_T(\cdot \mid c_t)\right) \\
&= \sum_{v \in S_t} P_S(v \mid c_t) \log \frac{P_S(v \mid c_t)}{P_T(v \mid c_t)}.
\end{split}
\end{equation}
Differentiating with respect to $z_j$ gives
\begin{equation}
\begin{split}
\frac{\partial \mathcal{L}^{\mathrm{RKL}}_t}{\partial z_j}
&= P_S(j \mid c_t)\Bigl(\log\frac{P_S(j \mid c_t)}{P_T(j \mid c_t)} \\
& - D_{\mathrm{KL}}\!\left(P_S(\cdot \mid c_t)\,\|\,P_T(\cdot \mid c_t)\right)\Bigr), \\
& \quad j \in S_t.
\end{split}
\label{eq:appendixreversegrad}
\end{equation}

The difference between Eq.~\ref{eq:appendixforwardgrad} and Eq.~\ref{eq:appendixreversegrad} is important in the intervention regime.
Forward KL yields a direct distribution gap correction on the selected support.
By contrast, the reverse KL gradient depends on the current student distribution in a more involved way.
Its update is explicitly scaled by $P_S(j \mid c_t)$, and the inner term is also determined by a student-dependent log ratio relative to the global reverse KL value.
As a result, the recovery signal is less direct when a teacher-preferred token has already been strongly suppressed by the student.
This comparison explains why forward KL is better suited to the teacher-defined support used in AOPD.

\subsection{Training Dynamics}
\label{sec:beta_training_dynamics}

We further examine the training dynamics under different $\beta$ values and observe an anomalous pattern when the objective tilts toward reverse KL. 
Figure~\ref{fig:beta_grad_norm} shows that as the divergence becomes more biased toward the student distribution, the optimization exhibits more unstable gradient behavior. 
In particular, the gradient norm exhibits the most severe spikes at $\beta=0$ and $0.1$, whereas the teacher-leaning settings $\beta=0.9$ and $1$ yield the smoothest training dynamics.
Figure~\ref{fig:beta_entropy} shows that the entropy under reverse KL ($\beta=0.0$) undergoes a sharp initial surge, followed by an abrupt collapse to near zero. 
This trajectory indicates that the student rapidly transits into a converged, low-entropy policy. 
Concurrently, Figure~\ref{fig:beta_token_ratio} reveals that the proportion of tokens receiving divergence guidance plummets at the same entropy collapse phase, implying that most generated positions register positive advantages and exit the intervention regime.
The ablation results in Section~\ref{sec:ablation_beta} demonstrate that this apparent convergence coincides with severe degradation in reasoning capability.
We attribute this to a reward-hacking phenomenon in on-policy distillation.
The student discovers a mode-collapsed strategy that artificially suppresses the reverse KL value while failing to acquire the teacher's reasoning patterns.
In contrast, forward KL maintains stable entropy and sustained intervention throughout training, yielding both superior and more robust performance.

\section{Implementation Details}
\label{sec:implementation_detail}

\paragraph{Environment.}
Experiments are implemented on top of VeRL~\citep{sheng2024hybridflow} (v0.3.1) for on-policy distillation training and vLLM~\citep{kwon2023efficient} (v0.8.5) for teacher model deployment.
Training is distributed via PyTorch FSDP and coordinated by Ray across 16 A100-80G GPUs, with an additional 8 A100-80G GPUs dedicated to serving the teacher model.

\begin{table}[h]
\centering
\small
\begin{tabular}{lc}
\toprule
\textbf{Hyperparameter} & \textbf{Value} \\
\midrule
\multicolumn{2}{l}{\textit{Training}} \\
Learning rate           & $1\times10^{-5}$ \\
Train batch size        & 512              \\
Max response length     & 12288            \\
Rollout temperature     & 1.0              \\
Teacher top-$K$         & 32               \\
\midrule
\multicolumn{2}{l}{\textit{Evaluation}} \\
Temperature             & 0.6              \\
top-$p$                 & 0.95             \\
top-$k$                 & 20               \\
Max generation length   & 32768            \\
\bottomrule
\end{tabular}
\caption{Training and evaluation parameters.}
\label{tab:hyperparams}
\end{table}

\paragraph{Parameters.}
Table~\ref{tab:hyperparams} summarizes the main training and evaluation parameters used in our experiments.
The full set of parameters is available in our \href{https://github.com/stargazer319/AOPD}{code repository}.

\section{Extended Experiments}

\subsection{Extended Main Results}
\label{sec:extended_main_result}
We further supplement Table~\ref{tab:qwen3_4b_1k_main} with the results of Qwen3-4B-Base under a 1K-step SFT warm-up initialization.
We observe that AOPD still attains the best overall performance, which is consistent with the trend observed in the main text.

\begin{table*}[htbp]
\centering
\small
\begin{tabular}{lccccccccc}
\toprule
\multicolumn{1}{c}{\multirow{2}{*}{\textbf{Method}}} & \multicolumn{2}{c}{\textbf{AIME 2024}} & \multicolumn{2}{c}{\textbf{AIME 2025}} & \multicolumn{2}{c}{\textbf{HMMT 2025 (Feb)}} & \multicolumn{2}{c}{\textbf{Average}} \\
\cmidrule(lr){2-3} \cmidrule(lr){4-5} \cmidrule(lr){6-7} \cmidrule(lr){8-9}
& \textit{Pass@1} & \textit{Pass@4} & \textit{Pass@1} & \textit{Pass@4} & \textit{Pass@1} & \textit{Pass@4} & \textit{Pass@1} & \textit{Pass@4} \\

\midrule
\multicolumn{9}{l}{\textbf{Qwen3-4B-Base}} \\
\quad \textit{-- 1k SFT Warm-up} \\
\quad \quad SFT Checkpoint   & 36.25 & 58.61 & 32.71 & 47.25 & 19.17 & 28.80 & 29.38 & 44.89 \\
\quad \quad + SeqKD          & 42.50 & 65.23 & 40.00 & 60.09 & 24.58 & 38.01 & 35.69 & 54.44 \\
\quad \quad + OPD            & 46.25 & 65.99 & 42.08 & 60.75 & 22.50 & 35.67 & 36.94 & 54.14 \\
\quad \quad + ExOPD          & 44.58 & 64.41 & 41.46 & 61.70 & 20.62 & 32.32 & 35.55 & 52.81 \\
\quad \quad + GKD            & \underline{50.21} & \underline{70.47} & \textbf{44.17} & \underline{63.03} & \underline{27.50} & \underline{42.21} & \underline{40.63} & \underline{58.57} \\
\quad \quad + \textbf{AOPD}  & \textbf{51.25} & \textbf{72.01} & \underline{43.96} & \textbf{63.49} & \textbf{28.33} & \textbf{46.02} & \textbf{41.18} & \textbf{60.51} \\
\bottomrule
\end{tabular}
\caption{
  Supplementary distillation results for Qwen3-4B-Base across multiple math reasoning benchmarks. We compare AOPD against SeqKD, OPD, GKD and ExOPD. \textbf{Bold} indicates the best result within each configuration block, while \underline{underlined} values denote the second best.
}
\label{tab:qwen3_4b_1k_main}
\end{table*}

\subsection{Detailed Training Curves}
\label{sec:detailed_training}

\begin{figure*}[h]
    \centering

    \begin{subfigure}[t]{0.85\textwidth}
        \centering
        \includegraphics[width=\textwidth]{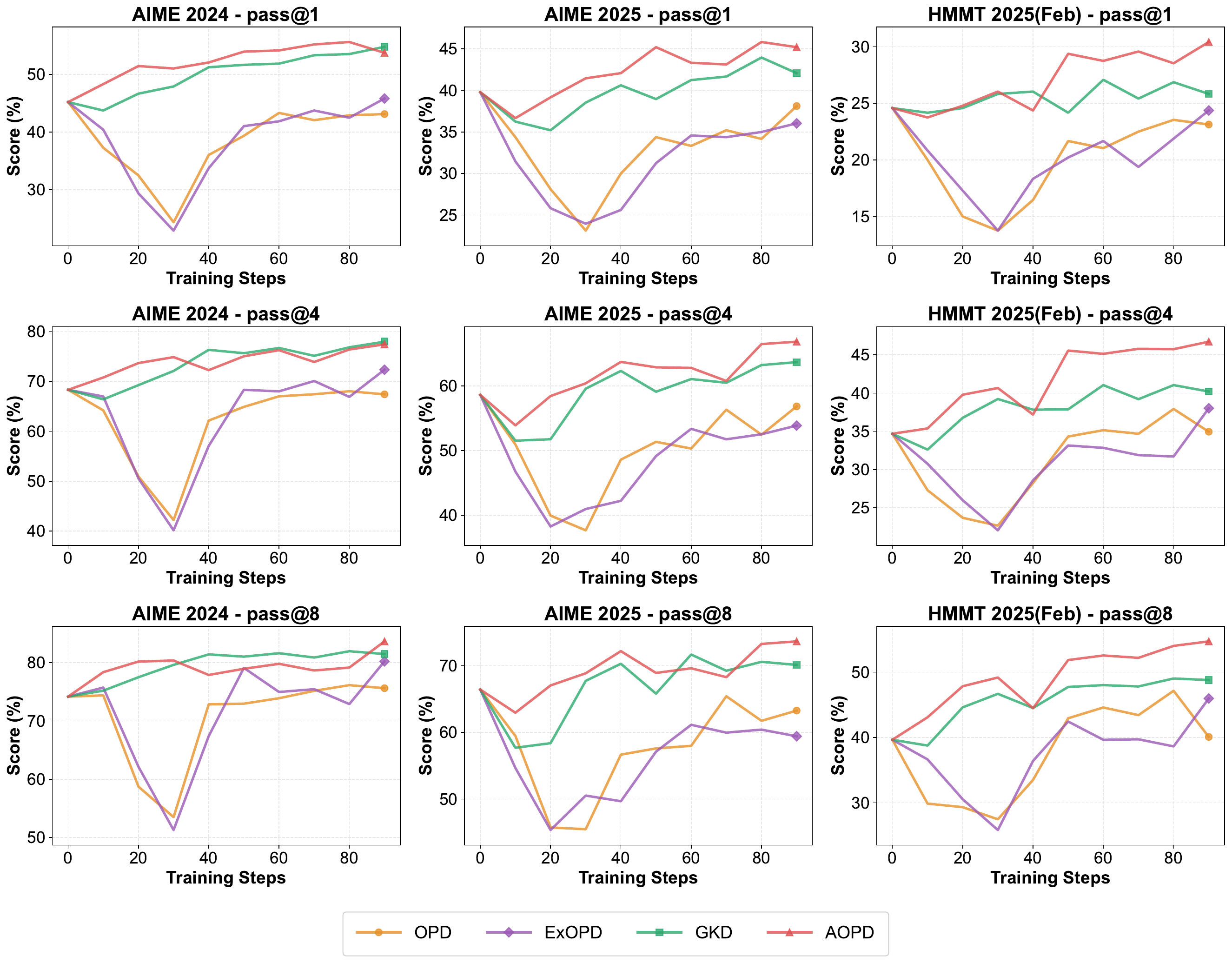}
        \caption{Weak initialization.}
        \label{fig:dynamics_sft1000}
    \end{subfigure}

    \vspace{0.5em}

    \begin{subfigure}[t]{0.85\textwidth}
        \centering
        \includegraphics[width=\textwidth]{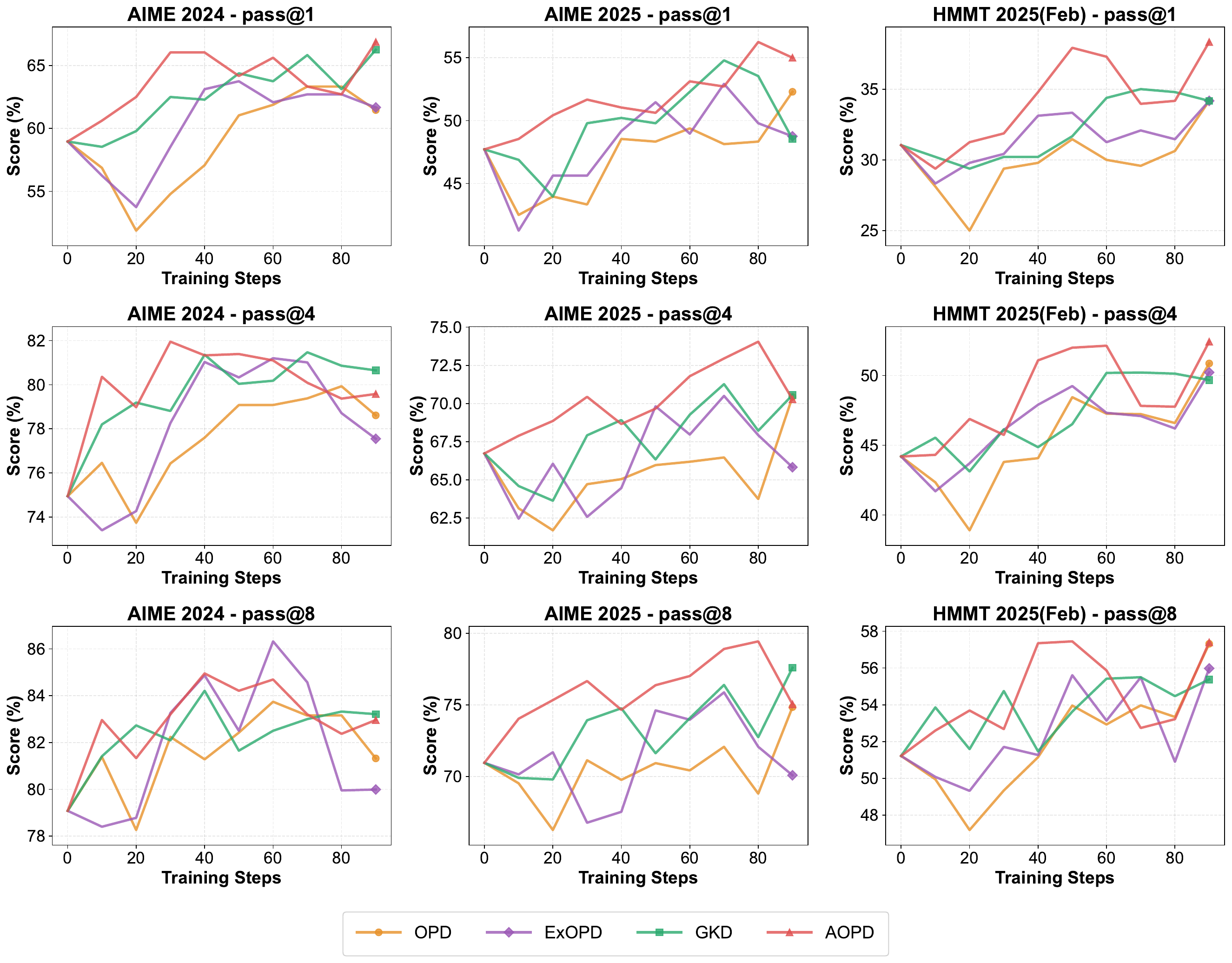}
        \caption{Strong initialization.}
        \label{fig:dynamics_sft3000}
    \end{subfigure}

    \caption{Detailed training dynamics of Qwen3-8B-Base under weak and strong initialization.}
    \label{fig:dynamics_both}
\end{figure*}

We report the full test-set performance dynamics throughout training for both warm-up settings on Qwen3-8B-Base. 
Figure~\ref{fig:dynamics_both} traces Pass@1/4/8 every 10 steps across the training trajectory.



Across both initialization settings, AOPD consistently achieves the highest scores on Pass@1/4/8 across all benchmarks. 
Under weak initialization, this advantage is particularly pronounced. 
The substantial performance gap indicates that AOPD remains effective even for students with weaker foundational capabilities. 
Under strong initialization, the margin between AOPD and baselines narrows, though AOPD still reaches the highest performance ceiling. 
This convergence suggests that methods such as OPD are more dependent on the student's initial capability. 
A stronger base model produces trajectories where student tokens more frequently receive positive advantages from the teacher, thereby reducing the frequency of exploration black holes and alleviating the structural weaknesses of negative reinforcement.

\subsection{Extended Ablation Studies on $\tau$}
\label{sec:ablation_tau}

\begin{figure}[h]
\centering
\begin{subfigure}[t]{\linewidth}
\centering
\includegraphics[width=0.9\linewidth]{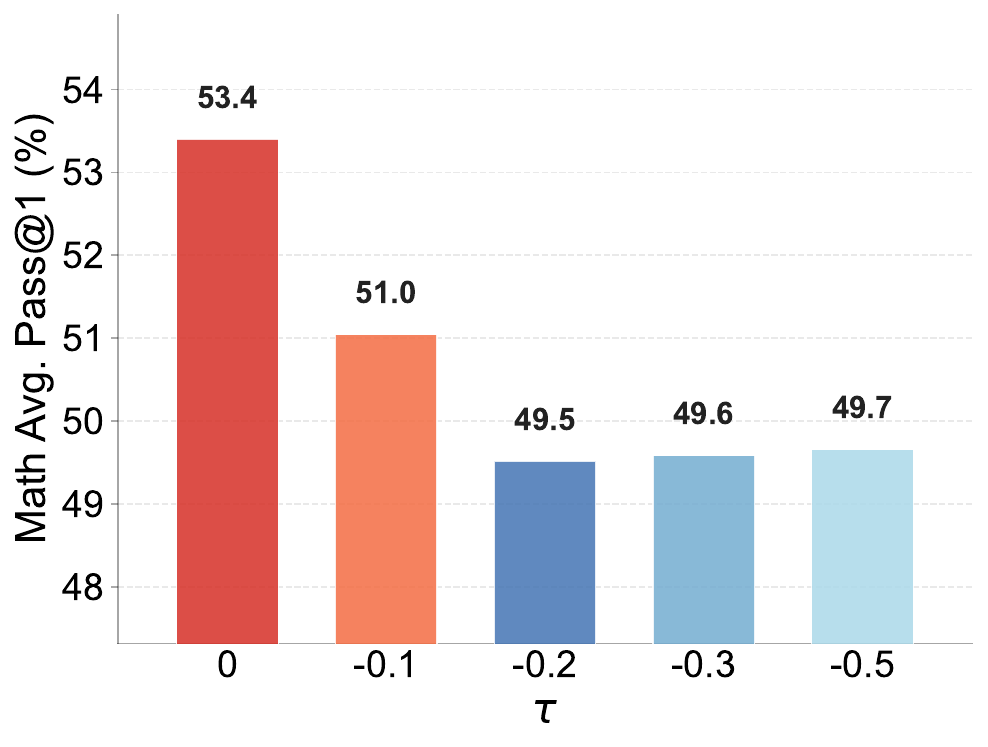}
\caption{Scores of different $\tau$}
\label{fig:tau_score}
\end{subfigure}

\begin{subfigure}[t]{\linewidth}
\centering
\includegraphics[width=0.9\linewidth]{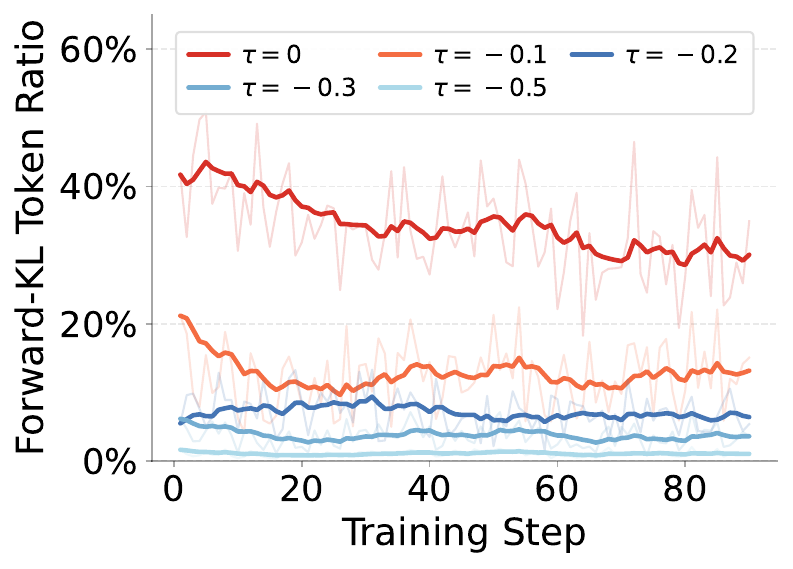}
\caption{Ratio of tokens receiving forward KL guidance.}
\label{fig:tau_token_ratio}
\end{subfigure}
\caption{Training dynamics under different $\tau$ values.}
\label{fig:tau_dynamics}
\end{figure}

We further conduct an extended study on the intervention threshold $\tau$.
As shown in Figure~\ref{fig:tau_score}, the mathematical reasoning performance gradually declines as $\tau$ decreases, and begins to plateau once $\tau$ reaches $-0.2$.
This trend suggests that reducing $\tau$ lowers the proportion of positions that receive forward KL supervision, thereby weakening the teacher-guided correction effect in AOPD.
As the intervention becomes increasingly sparse, the model correspondingly moves closer to standard OPD, and the performance eventually stabilizes.

This behavior is further explained by the intervention ratio in Figure~\ref{fig:tau_token_ratio}.
When $\tau \le -0.2$, the fraction of tokens that trigger the supervision loss drops to below 10\% of all generated tokens.
As a result, only a very small subset of positions receives explicit teacher guidance, which explains why the performance gap becomes relatively small once $\tau$ is reduced beyond $-0.2$.
These results suggest that the full benefit of AOPD is realized only when teacher guidance spans the entire zero-advantage and negative-advantage regions, whereas reducing forward KL supervision in the negative-advantage region correspondingly weakens the effect of AOPD.

\subsection{Comparisons with Other Methods}
\label{subsec:comparisons_with_other_methods}

\begin{table*}[h]
\centering
\small
\begin{tabular}{lcccccccc}
\toprule
\multirow{2}{*}{\textbf{Method}} & \multicolumn{2}{c}{\textbf{AIME 2024}} & \multicolumn{2}{c}{\textbf{AIME 2025}} & \multicolumn{2}{c}{\textbf{HMMT 2025 (Feb)}} & \multicolumn{2}{c}{\textbf{Average}} \\
\cmidrule(lr){2-3} \cmidrule(lr){4-5} \cmidrule(lr){6-7} \cmidrule(lr){8-9}
& \textit{Pass@1} & \textit{Pass@4} & \textit{Pass@1} & \textit{Pass@4} & \textit{Pass@1} & \textit{Pass@4} & \textit{Pass@1} & \textit{Pass@4} \\
\midrule
OPD & 61.88 & 79.08 & 49.38 & 66.19 & 30.00 & 47.26 & 47.09 & 64.18 \\
OPD-Norm & 60.62 & 77.80 & 50.21 & 68.49 & 32.92 & 46.73 & 47.92 & 64.34 \\
OPD-Clip & 62.08 & 78.81 & 51.46 & 67.50 & 33.96 & 50.29 & 49.17 & 65.53 \\
\midrule
KDRL & 64.58 & 80.62 & \textbf{53.54} & \textbf{72.30} & 36.46 & 51.11 & 51.53 & 68.01 \\
EOPD & \textbf{66.87} & 80.44 & 50.83 & 69.55 & 37.08 & 51.13 & 51.59 & 67.04 \\
\textbf{AOPD} & 65.62 & \textbf{81.10} & 53.12 & 71.79 & \textbf{37.29} & \textbf{52.13} & \textbf{52.01} & \textbf{68.34} \\
\bottomrule
\end{tabular}
\caption{Comparison with other stabilization methods, KDRL and EOPD. All methods are trained for 60 steps under the same experimental configuration. \textbf{Bold} indicates the best result in each column.}
\label{tab:compare_opd_variants}
\end{table*}

\paragraph{OPD Stabilization Methods.}
We compare AOPD with simpler strategies that stabilize the advantage-based update while retaining the original OPD optimization rule.
OPD-Norm normalizes token-level advantages along sequence dimension, while OPD-Clip limits their magnitudes in $[-1,1]$.
Both methods reduce the influence of high-variance advantages without changing the underlying policy-gradient update.
However, these stabilization strategies do not address the insufficient corrective directions in non-positive advantage regions, and therefore remain less effective than AOPD.

\paragraph{KDRL and EOPD.}
We further compare AOPD with methods that augment or adapt standard on-policy distillation.
KDRL~\citep{kdrl} combines distillation with outcome-level reinforcement learning signals, while EOPD~\citep{eopd} selectively introduces forward-KL guidance according to teacher entropy.
AOPD instead applies forward-KL guidance throughout the non-positive advantage region.
As shown in Table~\ref{tab:compare_opd_variants}, KDRL and EOPD also substantially improve over standard OPD.
AOPD achieves the best overall performance, reaching 52.01 average Pass@1 and 68.34 average Pass@4.
KDRL achieves comparable performance to AOPD, showing that additional outcome supervision can also effectively improve OPD.
Compared with EOPD, AOPD systematically covers the entire non-positive advantage region.
The stronger overall results suggest that more comprehensive coverage of these regions is beneficial for addressing the optimization bottlenecks of OPD.

\FloatBarrier

\section{LLM Usage Statement}
During the preparation of this manuscript, we utilized LLM as a general-purpose writing assistant. 
The use of LLMs was primarily for improving grammar, spelling, and clarity.
We reviewed and edited all suggested changes and take full responsibility for the final content of this paper.
All research contributions, including problem formulation, methodology design, implementation, experiments, data analysis, and scientific conclusions, were conceived, executed, and verified by the authors. 
This use conforms to the ACL Ethics Policy regarding authorship criteria.

\end{document}